\documentclass[lettersize,journal,twoside]{IEEEtran}
\usepackage{amsmath,amsfonts}
\usepackage{algorithmic}
\usepackage{algorithm}
\usepackage{array}
\usepackage{amsmath}
\usepackage{amssymb}
\usepackage{pifont}
\usepackage[caption=false,font=normalsize,labelfont=sf,textfont=sf]{subfig}
\usepackage{textcomp}
\usepackage{stfloats}
\usepackage{url}
\usepackage{verbatim}
\usepackage[dvips]{graphicx}
\usepackage[dvips]{epsfig}
\usepackage{cite}
\usepackage{booktabs}
\usepackage{multirow}
\usepackage{svg}
\usepackage{caption}
\newcolumntype{C}[1]{>{\centering\let\newline\\\arraybackslash\hspace{0pt}}m{#1}}
\newcommand{\cmark}{\ding{51}}
\usepackage[capitalize]{cleveref}
\crefname{section}{Sec.}{Secs.}
\Crefname{section}{Section}{Sections}
\Crefname{table}{Table}{Tables}
\crefname{table}{Tab.}{Tabs.}

\begin{document}

\title{Cross-aware Early Fusion with Stage-divided Vision and Language Transformer Encoders for Referring Image Segmentation}

\author{Yubin Cho\thanks{Yubin Cho and Hyunwoo Yu contributed equally to this work. (Corresponding author: Suk-Ju Kang.)}, Hyunwoo Yu, and Suk-Ju Kang,~\IEEEmembership{Member,~IEEE}
\thanks{Yubin Cho is with the School of Artificial Intelligence, Sogang University, Seoul, 04017, Republic of Korea (e-mail: dbqls1219@sogang.ac.kr).}
\thanks{
Hyunwoo Yu, and Suk-Ju Kang are with the School of Electronic Engineering, Sogang University, Seoul, 04017, Republic of Korea (e-mail: hyunwoo137@sogang.ac.kr; sjkang@sogang.ac.kr).}
}

\markboth{IEEE TRANSACTIONS ON MULTIMEDIA, VOL. 26, 2024}%
{CHO \MakeLowercase{\textit{et al.}}: Cross-aware early fusion with stage-divided vision and language transformer encoders}


\maketitle

\begin{abstract}
Referring segmentation aims to segment a target object related to a natural language expression. Key challenges of this task are understanding the meaning of complex and ambiguous language expressions and determining the relevant regions in the image with multiple objects by referring to the expression. Recent models have focused on the early fusion with the language features at the intermediate stage of the vision encoder, but these approaches have a limitation that the language features cannot refer to the visual information. To address this issue, this paper proposes a novel architecture, Cross-aware early fusion with stage-divided Vision and Language Transformer encoders (CrossVLT), which allows both language and vision encoders to perform the early fusion for improving the ability of the cross-modal context modeling. Unlike previous methods, our method enables the vision and language features to refer to each other's information at each stage to mutually enhance the robustness of both encoders. Furthermore, unlike the conventional scheme that relies solely on the high-level features for the cross-modal alignment, we introduce a feature-based alignment scheme that enables the low-level to high-level features of the vision and language encoders to engage in the cross-modal alignment. By aligning the intermediate cross-modal features in all encoder stages, this scheme leads to effective cross-modal fusion. In this way, the proposed approach is simple but effective for referring image segmentation, and it outperforms the previous state-of-the-art methods on three public benchmarks.
\end{abstract}

\begin{figure}[t] 
\includegraphics[width=\linewidth]{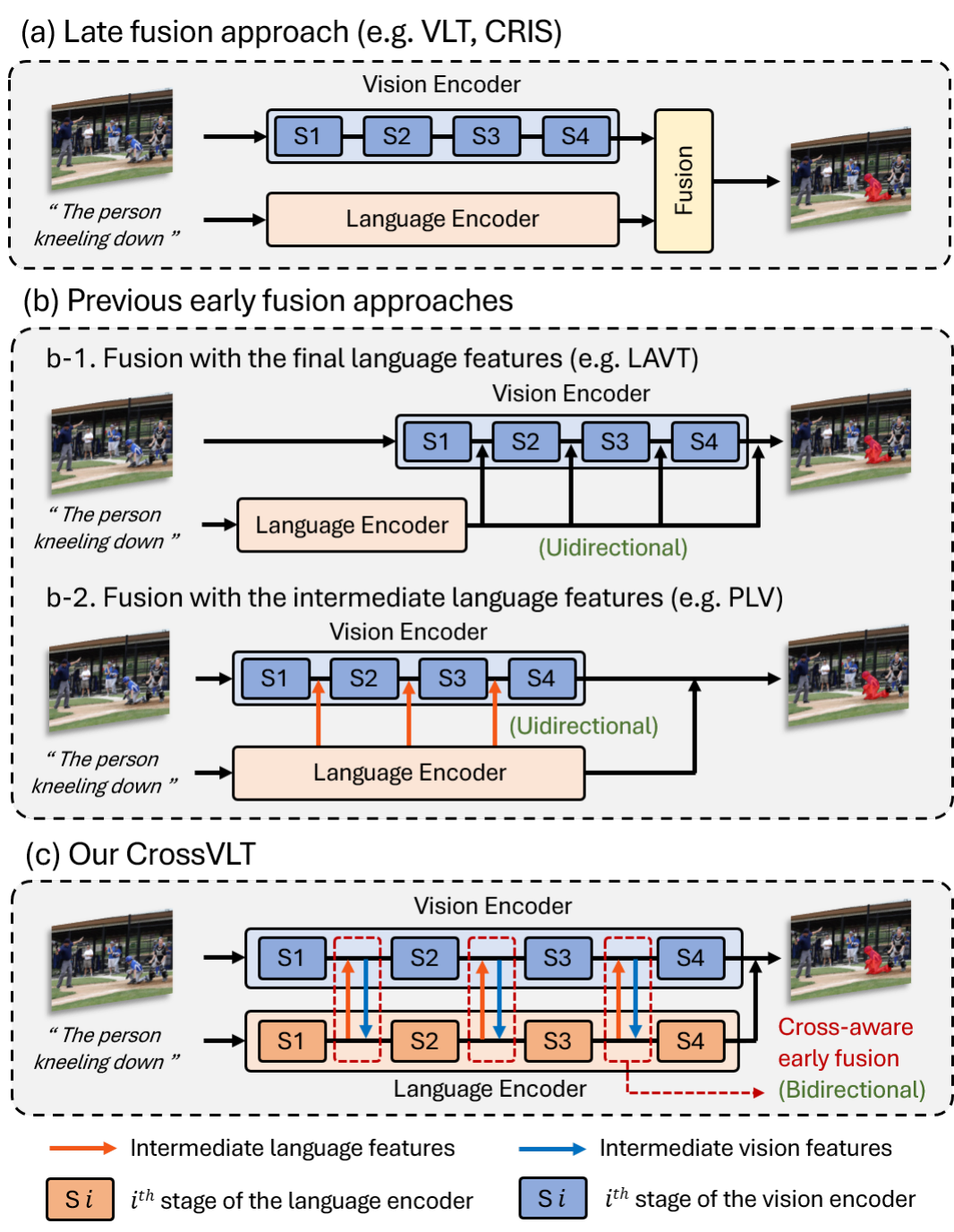}
\caption{Architectures of various fusion approaches for referring image segmentation. (a) Late fusion approach (e.g., VLT\cite{ding2021vision}, CRIS\cite{wang2022cris}) that fuses in the transformer decoder after the encoder feature extraction. (b) Previous early fusion approaches (e.g., LAVT\cite{yang2022lavt}, PLV\cite{liao2022progressive}) that unidirectionally refer to the language features in the vision encoder. (c) Our CrossVLT that bidirectionally performs cross-aware early fusion at each stage to interconnect both encoders for mutual enhancement.}
\label{fig:fig1}
\end{figure}

\section{Introduction}
\IEEEPARstart{R}{eferring} image segmentation \cite{hu2016segmentation,Alpher01,ye2019cross, qiu2019referring, shi2020query, ye2020dual, lin2021structured, hua2023multiple} is one of the most fundamental and challenging vision-language tasks to highlight regions corresponding to the language description for the properties of the target object. It can be applied in various applications such as human-robot interaction and image editing. Unlike the conventional single-modal segmentation (e.g., instance or semantic segmentation \cite{shuai2018toward,zhou2019semantic, ding2020semantic, xie2021segformer, guo2022segnext, shim2023feedformer}) based on fixed category conditions, referring image segmentation aims to determine the region according to various linguistic expressions that are not predetermined. In this task, the images with dense objects have complicated relationships between the target object and other objects. In addition, linguistic expressions are ambiguous because the linguistic meaning can be interpreted in various ways depending on the perspectives. Thus, the context of the image should be taken into account to interpret the linguistic expression with the appropriate meaning. To resolve these problems, we address two main challenges in the referring image segmentation task. One is regarding the fusion that refers to the cross-modal information from each modality features. The other challenge is the alignment that embeds the vision and language features into the same space to improve the fusion performance \cite{yang2022vision}.

The first challenging point, \emph{cross-modal fusion}, makes it possible to refer to the mutual information between an image and a language expression. Several fusion approaches have been proposed for this task. The fusion approaches can be categorized into \emph{late fusion} and \emph{early fusion}. As shown in \cref{fig:fig1}(a), most typical late fusion approach \cite{ding2021vision,wang2022cris,kim2022restr} leverages the decoder architectures that fuse the final features extracted from the vision and language encoders. This approach, which depends only on the decoder architectures for the cross-modal fusion, cannot effectively utilize the rich encoder information. Recently, the early fusion approach \cite{feng2021encoder,yang2022lavt, liao2022progressive} performs the fusion using the vision features as a query and the language features as a key-value in the intermediate stages of the vision encoder. We call this early fusion approach as vision-only early fusion. As illustrated in \cref{fig:fig1} (b-1), previous early fusion methods \cite{feng2021encoder, yang2022lavt} perform the vision-only early fusion by passing the final language features to each vision stage. Another vision-only early fusion method \cite{liao2022progressive} provides the intermediate language encoder features to each vision stage, as described in \cref{fig:fig1} (b-2). By achieving superior improvements, these early fusion methods demonstrate that the vision-only early fusion approach is more effective than the late fusion approach for the cross-modal interactions. However, due to unidirectional early fusion, the language encoder cannot utilize the visual information. Therefore, the vision-only early fusion approach \cite{yang2022lavt, liao2022progressive} has a limitation in interpreting the unrestricted language expressions into the appropriate meaning that considers the context of the image in this task.

The second challenging point, \emph{alignment}, is to embed the cross-modal features of each encoder into the same embedding space for effective vision-language feature fusion. To perform cross-modal alignment, vision-language pretraining tasks \cite{radford2021learning,jia2021scaling,li2021align} use the text-to-image contrastive learning that embeds positive pairs closely and negative pairs distantly. Specially, \cite{li2021align} demonstrates that aligning the cross-modal features before performing the fusion is beneficial for the fusion. To apply the contrastive learning to the pixel-level prediction task, the text-to-pixel contrastive learning \cite{wang2022cris} based on the conventional text-to-image method is employed to promote the relation between two modalities at the pixel-level. This scheme has improved the performance of the vision-language alignment in this task. However, these approaches use only the final features of the vision and language encoders or the output logits to align the cross-modal features.  Therefore, the intermediate features of the vision and language encoders, which have rich information where low-level features contain the structural (or syntactic) information and high-level features contain the semantic global information \cite{tenney2019bert,peters2018dissecting}, cannot be sufficiently aligned to the joint embedding space.

This paper proposes a novel architecture for referring image segmentation, Cross-aware early fusion with stage-divided Vision and Language Transformer encoders (CrossVLT), which mutually enhances the robustness of each encoder by extending the early fusion approach to a language encoder as well as a vision encoder. Typically, the vision encoder consists of the multiple stages that are separated by the resolution of the feature map; such a structure, which is divided into multiple stages based on specific criteria (e.g., pooling the feature resolution), is called a \textit{stage-divided} structure in this paper. Different from the vision encoders, the language encoders are not commonly designed with the stage-divided structure. To enable to bidirectionally exchange the cross-modal information of two encoders at each stage, our CrossVLT is designed with the stage-divided language and vision encoders as shown in \cref{fig:fig1} (c). Unlike previous methods \cite{yang2022lavt, liao2022progressive}, our language encoder considering the visual contexts provides more contextual language features to the vision stages. Our vision and language encoders can exchange richer information at each stage and jointly improve their ability to understand the context of images and expressions by considering each other's perspectives. Therefore, our method advances the ability of the cross-modal context modeling to deal with the complicated images and the ambiguous language expressions. In addition, unlike the conventional scheme that relies solely on the final features for the cross-modal alignment, we introduce a feature-based alignment scheme where the low-level to high-level features participate in the cross-modal alignment. Prior to performing the fusion, our scheme can sufficiently align the intermediate cross-modal features to the joint embedding space, leading to a more effective fusion at each stage. We also employ the alignment loss to circumvent the conflict with a task loss. Our scheme better allows the vision and language encoders to learn the cross-modal correlation by improving the ability of aligning the cross-modal features.

We demonstrate the effectiveness of the proposed method by achieving the competitive performance on three public datasets for referring image segmentation. In addition, we empirically divide the stages of the language encoder based on the optimal position where the cross-modal alignment and cross-aware fusion with the vision features of the corresponding vision stage can be effectively performed.
Our contributions are summarized as follows.
\begin{enumerate}
    \item We propose a novel network for referring image segmentation, CrossVLT, which leverages the stage-divided vision and language transformer encoders to perform the cross-aware early fusion and mutually enhances the robustness of each encoder.
    \item We introduce a feature-based alignment scheme that enables the low-level to high-level features of the vision and language encoders to engage in the cross-modal alignment to improve the ability of aligning the intermediate cross-modal features. Our scheme leads to a more effective cross-modal fusion at each stage by being applied prior to performing the fusion.
    \item Our CrossVLT is simple yet effective, and outperforms the previous state-of-the-art methods on three widely used datasets for referring image segmentation.
\end{enumerate}

\section{Related Works}
\subsection{Fusion for Referring Image Segmentation}
Different from the conventional segmentation task \cite{ shuai2018toward,zhou2019semantic,ding2020semantic, xie2021segformer, guo2022segnext, shim2023feedformer} based on fixed categories, referring image segmentation aims to find the target object according to the unrestricted language expressions. In the referring image segmentation task, various methods \cite{hu2016segmentation,li2018referring, ye2019cross,huang2020referring, ye2020dual, liu2022instance} have been introduced to fuse vision and language features. Hu \textit{et al.} \cite{hu2016segmentation} and RRN \cite{li2018referring} concatenated language features encoded by LSTM \cite{sak2014long} and vision features encoded by CNNs \cite{krizhevsky2012imagenet} to generate the fused features. CMSA \cite{ye2019cross} also concatenated each feature extracted from vision and language encoders, then used the self-attention to captures the long-range dependencies between visual and linguistic features. CMPC \cite{huang2020referring} adopted the bilinear fusion to associate spatial regions with correlated linguistic features of the entity and attribute words. To improve the multi-modal interaction towards more important words, Ye et al. \cite{ye2020dual} proposed a multi-modal feature encoder that fuses the visual and linguistic features by using a dual convolution LSTM framework. Liu \textit{et al.} \cite{liu2022instance} used the element-wise multiplication for the fusion.

\begin{figure*}[t] 
\centering
\includegraphics[width=0.98\linewidth]{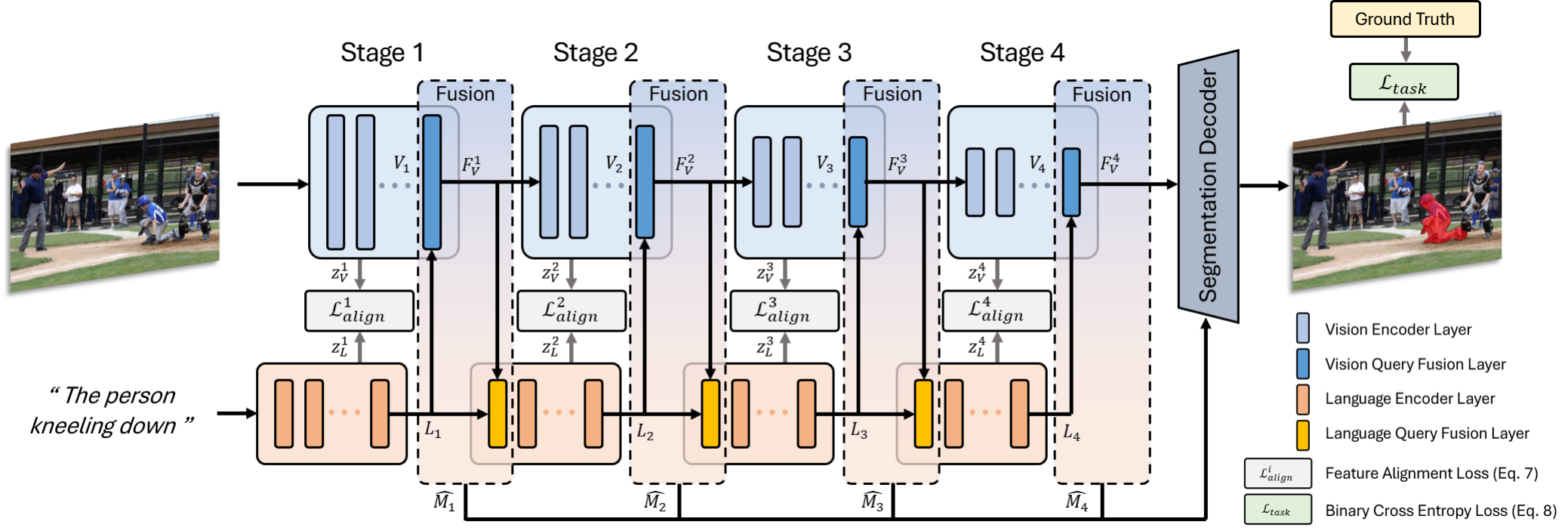}
\caption{Overview of CrossVLT, consisting of the stage-divided vision and language encoders, the feature-based alignment, and the segmentation decoder. At each stage, the vision and language encoders consider each other’s features through cross-aware fusion to capture the rich contextual information in each encoder. The feature-based alignment is used to better embed the vision and language features into the same space by applying the contrastive learning to the intermediate stages of each encoder.}
\label{fig:fig2}
\end{figure*} 

Recent methods \cite{ding2021vision, zhu2022seqtr, kim2022restr, wang2022cris} have used a cross-modal fusion module after extracting features from the uni-modal encoders. VLT \cite{ding2021vision} adopted a transformer encoder-decoder structure that extracts vision features in the encoder and fuses vision and language features in the transformer decoder. SeqTR \cite{zhu2022seqtr} also utilized the transformer encoder-decoder structure after the feature extraction and fusion. ReSTR \cite{kim2022restr} used transformer-based feature extractor and the visual-linguistic transformer encoder. CRIS \cite{wang2022cris} extracted the encoder features from the image encoder and language encoder of CLIP \cite{radford2021learning} and fused them in a vision-language decoder. 

For better cross-modal fusion, recent methods \cite{feng2021encoder, yang2022lavt, liao2022progressive} performed the early fusion in the vision encoder stages instead of fusing after feature extraction. EFN \cite{feng2021encoder} conducted the cross-attention using linguistic features as a key-value in the intermediate stages of the vision encoder. LAVT \cite{yang2022lavt} also performed the language-aware visual attention using the final language features of BERT \cite{devlin2019bert} as a key-value on all stages of Swin \cite{liu2021swin}. PLV \cite{liao2022progressive} provided the intermediate language features to the vision stages for the language-aware visual attention. These studies have demonstrated the effectiveness of the vision-only early fusion, but the enhancement of the language encoder was not considered. Unlike previous methods, we propose the cross-aware early fusion with stage-divided vision and language transformer encoders. In our method, both vision and language encoders jointly perform the early fusion to better understand expressions for accurate segmentation.

\subsection{Alignment for Vision-Language Task}
Contrastive learning, which exploits positive pairs closely and negative pairs distantly, has been effectively used in vision-language tasks for alignment. In vision-language pretraining tasks, CLIP \cite{radford2021learning} and ALIGN \cite{jia2021scaling} applied the vision-language contrastive loss for cross-modal matching using massive web data that consists of image-text pairs. In addition, ALBEF \cite{li2021align} proposed the align-before fusion framework that applies the image-text contrastive loss to the cross-modal features of the final encoder layers before fusing them. The align-before fusion framework demonstrated that the cross-modal features alignment is enhanced by considering the intermediate vision-language features. In referring image segmentation, CRIS \cite{wang2022cris} performed text-to-pixel contrastive learning using the knowledge distillation from CLIP \cite{radford2021learning} that applies the text-to-image contrastive learning.

Different from previous methods, we propose the feature-based alignment using contrastive loss based on the features of the intermediate stages in each encoder. It leverages the text-to-pixel contrastive learning and the align-before fusion. However, to the best of our knowledge, this is the first attempt that considers the features of the intermediate stages for cross-modal alignment in referring image segmentation. We also explored the appropriate design of the loss that can be applied to the intermediate layers for better alignment without conflicting with the task loss.

\section{METHODOLOGY}

As illustrated in \cref{fig:fig2}, we introduce CrossVLT, which is designed to mutually enhance the robustness of vision and language encoders and improve the ability of cross-modal feature alignments. First, from low-level to high-level stages, each stage of the two encoders extracts vision and language features, and these features are effectively mapped to the cross-modal embedding space by performing feature-based alignment in the intermediate layers of the two encoders. Then, the cross-aware early fusion is performed to enrich the contextual information of each modality. Finally, the simple segmentation decoder utilizes vision encoder features from each fusion block to generate a output mask for a target object.

\subsection{Stage-divided Vision and Language Encoders}
CrossVLT is designed with a stage-divided language encoder as well as a stage-divided vision encoder to mutually enhance the robustness of both encoders by jointly performing early fusion. Given an image-text pair, the vision stage takes an image as an input and the language stage takes a language expression as an input. Each stage is indexed as ${i} = 1,2,\cdots,{n}.$

\textbf{Stage-divided vision encoder.}
For the dense prediction task, we adopt a Swin transformer \cite{liu2021swin} organized into four stages to capture long-range visual dependencies and extract hierarchical features defined as ${V_{i}\in \mathbb{R}_{}^{H_{i}W_{i}\times C_{i}}}$. We append a vision query fusion layer as the last layer of each vision stage, which consists of the multi-head cross-attention and feed-forward block for fusion with language features. Note that ${H_{i}}$,${W_{i}}$, and ${C_{i}}$ denote the height, width, and channel dimension of the feature maps at the ${i^{th}}$ vision stage.

\textbf{Stage-divided language encoder.}
Unlike typical language encoders \cite{vaswani2017attention,devlin2019bert} without dividing stages, we newly design a stage-divided language encoder based on the BERT-base model \cite{devlin2019bert}, which is a representative transformer-based language encoder, to jointly perform cross-aware early fusion with the vision encoder. The stage-divided language encoder is divided into four stages to correspond to each stage of the vision encoder. Unlike the vision encoder, where stages are distinctly divided according to resolution, the language encoder has no standard to divide stages clearly. Empirically, the number of layers in each stage is set to [6,2,2,2], which can sufficiently extract the linguistic information at low-level. For more details, please refer to \Cref{tab:tableA1}. In the 2nd to 4th language stages, we replace self-attention mechanism of the first encoder layer with the language query cross-attention mechanism for fusion with vision features. Each language stage extracts the language features $\mathit{L_i \in \mathbb{R}^{T\times D}}$, where $T$ and $D$ denote the length of the expression and the feature dimension. The first token of the language features is a special \verb|[CLS]| token that is encoded to include representative information of all language tokens for understanding at sentence-level. The \verb|[CLS]| token of $\mathit{L_i}$ is defined as $\mathit{CLS_i \in \mathbb{R}^{1\times D}}$.

\subsection{Cross-aware Early Fusion}
\begin{figure}[t]
\includegraphics[width=1.0\linewidth]{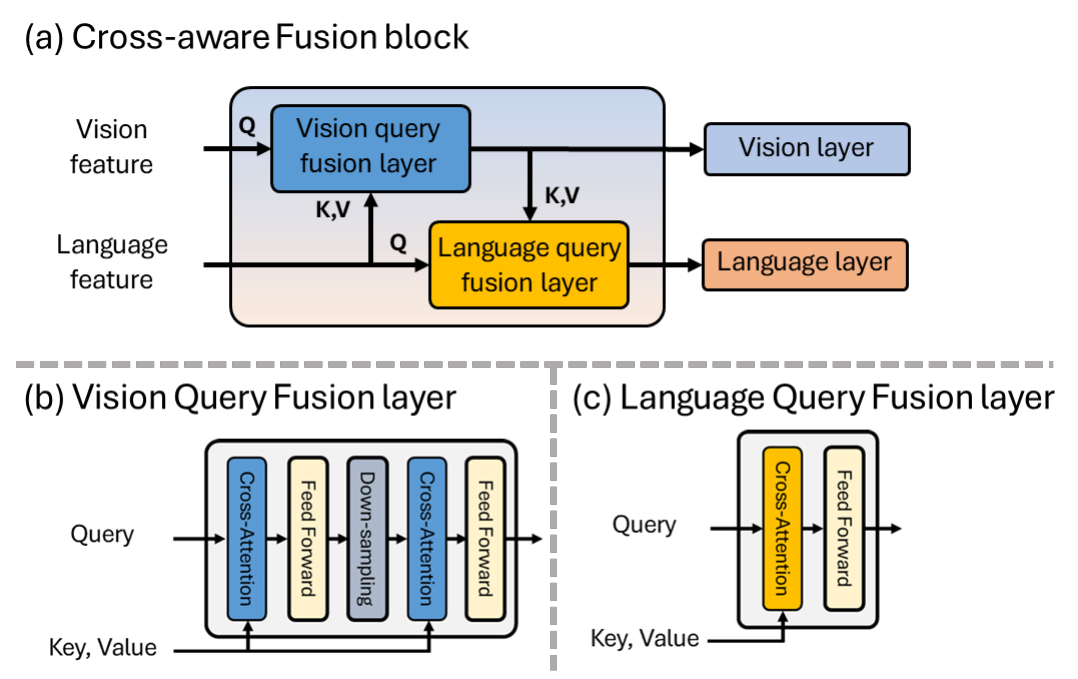}
\caption{(a) The cross-aware fusion block fuses the cross-modal information bidirectionally. (b) The vision query fusion layer consists of two cross attentions with downsampling to consider language-aware multi-scale vision features. (c) The fusion layer using language features as a query.}
\label{fig:fig3}
\end{figure}

The cross-aware fusion block fuses the vision and language features by traversing through each stage of both encoders alternately to extract robust features considering the perspective of the other modality. The contextual information of the vision features is enriched by referring to the linguistic information relevant to each pixel of the vision feature maps. As presented in \cref{fig:fig3}, the vision query fusion layer of the $\mathit{i^{th}}$ vision stage takes the language features $\mathit{L_i}$ and the vision features $\mathit{V_i}$ as inputs. The fusion layer performs the multi-head cross attention using vision features as queries and language features as key-value pairs. Then, the feed-forward block extracts the language-aware vision features $\mathit{M_{i}\in \mathbb{R}_{}^{H_{i}W_{i}\times C_{i}}}$. Specifically, the fusion process of the vision stage is described as follows. 
\begin{equation}
\mathit{MHCA}(\mathit{Q},\mathit{K},\mathit{V})=\mathrm{Softmax}(\mathit{Q}\cdot\mathit{K}^\mathit{T}/\sqrt{d_k})\cdot\mathit{V},
\end{equation}
\begin{equation}
\mathit{\widehat{M}_i = MHCA(V_i, L_i)} + \mathit{V_i}\ ,\ \mathit{M_i} = \mathit{FFN(\widehat{M}_i)}+\mathit{V_i} \ ,
\end{equation}
\begin{equation}
\mathit{D_i = \mathrm{Down}(M_i)\ , \ \widehat{F}^i_V = MHCA(D_i, L_i) + D_i} \ ,
\end{equation}
\begin{equation}
\mathit{F^i_V = FFN(\widehat{F}^i_V)+D_i} \ ,
\end{equation}
where $ \mathit{Q,K,V}$ and $ \mathit{d_k}$ denote queries, keys, values and dimensions of keys. $ \mathit{MHCA(\cdot)}$ and $ \mathit{FFN(\cdot)}$ indicate the multi-head cross attention and the feed-forward block. $ \mathrm{Down(\cdot)}$ denotes $\frac{1}{2}\times$downsampling. The intermediate vision features $\mathit{\widehat{M}_i}\in\mathbb{R}_{}^{{H_{i}\times W_{i}\times C_{i}}}$ are used for the segmentation decoder and the language-aware vision features $\mathit{M_i}$ are downsampled except for the final stage. The downsampled vision features evolve into the language-aware vision features $\mathit{F^i_V}\in\mathbb{R}_{}^{{H_{i+1}W_{i+1}\times C_{i+1}}}$ through the multi-head cross attention and feed-forward. Then, $\mathit{F^i_V}$ is passed to the next vision stage and the language query fusion layer of the next language stage. 

Afterward, the language query fusion layer of the $\mathit{(i+1)^{th}}$ language stage takes the vision features $\mathit{F^i_V}$ and language features $\mathit{L_i}$ as inputs. The multi-head cross attention is performed in this fusion layer, which uses the language features as queries and vision features as key-value pairs to understand the linguistic meanings from visual perspectives. Then, the vision-aware language features $\mathit{F^{i}_L\in\mathbb{R}^{T\times D}}$ are extracted by the feed-forward block and residual connections. The fusion process of the language stage is described as follows:
\begin{equation}
\mathit{\widehat{F}^i_L = MHCA(L_i, F^i_V)} + \mathit{L_i}\ ,\ \mathit{F^i_L} = \mathit{FFN(\widehat{F}^i_L)}+\mathit{\widehat{F}^i_L} \ ,
\end{equation}
where $\mathit{\widehat{F}^i_L}$ denotes the intermediate language features. The vision-aware language features $\mathit{F^i_L}$ are passed to the subsequent language layers.

\subsection{Feature-based Alignment}
\begin{figure}[t] 
\includegraphics[width=1.0\linewidth]{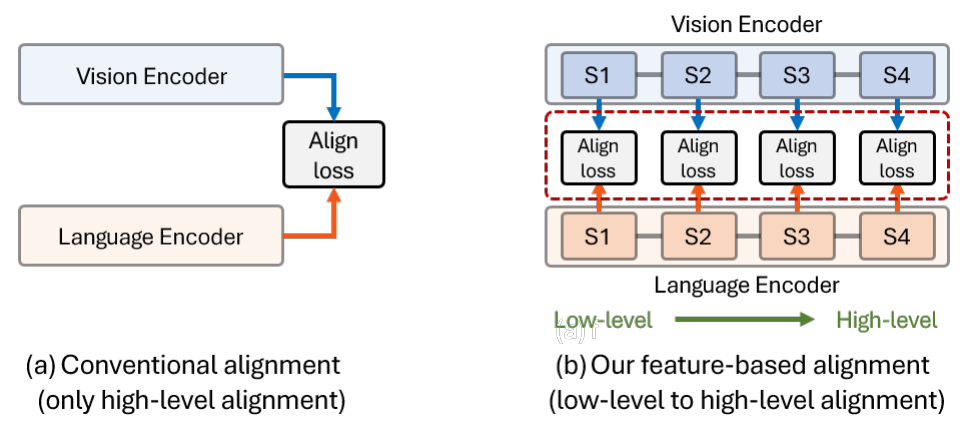}
\caption{{The structure of the conventional alignment and our feature-based alignment. (a) The final features are solely responsible for aligning the vision and language features. (b) The low-level to high-level features engage in the alignment for a more comprehensive alignment of the intermediate cross-modal features.}}
\label{fig:align}
\end{figure}

In most previous methods \cite{li2021align,radford2021learning,wang2022cris}, the final features of the vision and language encoders are solely responsible for the cross-modal feature alignment as illustrated in Fig. \ref{fig:align}. Unlike these methods, our feature-based alignment scheme engages the low-level to high-level features in the cross-modal alignment to more effectively embed the intermediate features of the vision and language encoders into the cross-modal embedding space. We now explain the position where the features are aligned for the effective cross-modal fusion and describe how to compute the alignment loss.

\textbf{Feature alignment position.} The position of the alignment is in front of the cross-aware fusion block at each stage. That is, the vision features $\mathit{V_i}$ and language features $\mathit{L_i}$ are adopted to conduct the feature-based alignment. This positioning makes it better for the vision and language encoders to learn the cross-modal correlation by grounding the vision and language features used for the fusion.

\begin{figure}[t] 
\includegraphics[width=1.0\linewidth]{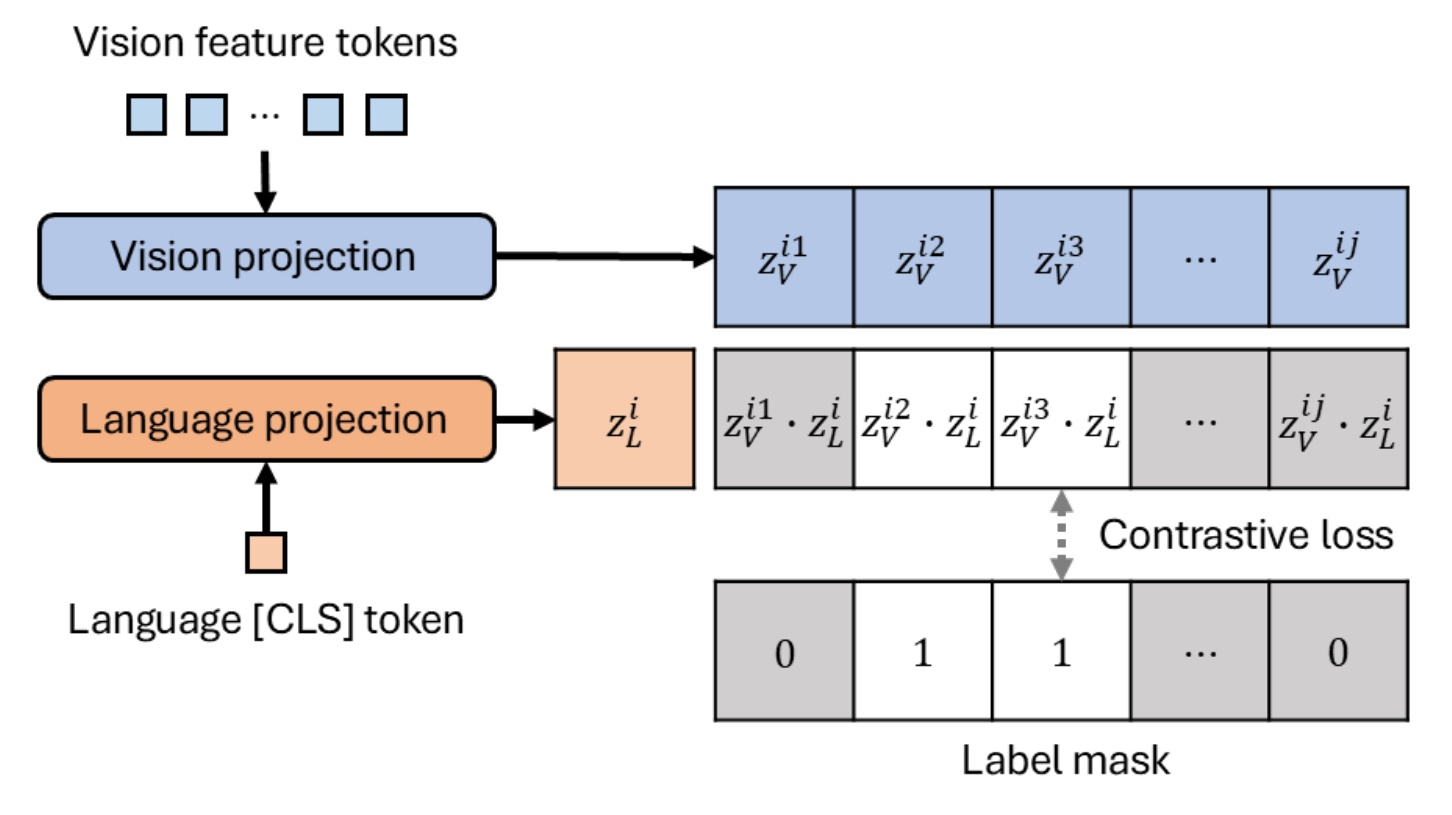}
\caption{The scheme of the alignment loss using vision feature tokens and a language [CLS] token to embed the cross-modal features into the same space.}
\label{fig:fig4}
\end{figure}

\textbf{Feature alignment loss.} We use the text-to-pixel contrastive loss as an alignment loss for this pixel-level prediction task. As illustrated in \cref{fig:fig4}, the vision features $\mathit{V_{i}}$ and \verb|[CLS]| token of language features $\mathit{CLS_i}$ are used to compute the text-to-pixel contrastive loss ${\mathcal{L}_{align}}$. Vision and language linear projections transform $\mathit{V_i}$ and $\mathit{CLS_{i}}$ into the same feature dimension $\mathit{D^{'}}$, respectively. The transformed vision features $\mathit{z^{i}_{V}\in \mathbb{R}_{}^{H_{i}W_{i}\times D^{'}}}$ and the transformed language features $\mathit{z^{i}_{L}\in \mathbb{R}_{}^{1\times D^{'}}}$ are used to obtain a similarity map. The feature alignment loss is calculated as follows:
\begin{equation}
{\mathcal{L}^{ij}_{align}}=\begin{cases}-\mathrm{log}(\sigma (\mathrm{Sim}(z^{ij}_{V},z^{i}_{L})/\tau _{i}))&j\in Z^{+}\\
-\mathrm{log}(1-\sigma (\mathrm{Sim}(z^{ij}_{V},z^{i}_{L})/\tau _{i}))&j\in Z^{-} ,\end{cases}
\end{equation}
\begin{equation}
\mathcal{L}_{align}=\frac{1}{|Z|}\sum^{n}_{i=1}\sum_{j\in Z}\mathcal{L}^{ij}_{align}\ ,
\end{equation}
where $Z$, $Z^+$ and $Z^-$ denote the set of pixels, relevant pixels (positive) and irrelevant pixels (negative) for language expression, $\mathrm{Sim}$ is a cosine similarity, $\tau$ is learnable temperature parameters, and $\sigma$ is a sigmoid function. We leveraged a sigmoid function with learnable parameters for loss calculation to circumvent the conflict with a task loss. $\mathcal{L}_{align}$ optimizes the networks so that relevant pixels are embedded close to the language features and irrelevant pixels are embedded far apart.

\subsection{Segmentation Decoder}
The segmentation decoder is designed with a simple structure, where three decoder blocks are stacked to verify the effectiveness of our encoders. The decoder block consists of two layers stacked using $3\times3$ convolution, batch normalization and a ReLU function. The decoder features are upsampled using the bilinear interpolation, and fed into the next decoder block after concatenating them with the vision features $\widehat{M}_{i}$ extracted by vision query cross attention module of the $i^{th}$ vision stage. The final prediction mask is projected into a binary class mask by conducting a $1\times1$ convolution. We use the binary cross-entropy loss $\mathcal{L}_{task}$ for network training. Thus, the final loss function is as follows.
\begin{equation}
\mathcal{L}_{task}=-\frac{1}{N}\sum^{N}_{j=1}{[y_j\mathrm{log}(p_j)+(1-y_j)\mathrm{log}(1-p_j)]}\ ,
\end{equation}
\begin{equation}
\mathcal{L}_{total}=\mathcal{L}_{task}+\lambda\cdot \mathcal{L}_{align}\ ,
\end{equation}
where $N, y_j$ and $p_j$ denote the total number of the pixels, the truth value and the predicted probability for the $j^{th}$ pixel.

\begin{table*}[t]
\centering
\renewcommand{\arraystretch}{1.1}\caption{Performance comparison with previous state-of-the-art methods using oIoU (\%). Models used the transformer structure have a \cmark\ mark . $^{\dagger}$ \ indicates that the code or pretrained model is not available. The best results are in \textbf{bold}, and the second-best results are \underline{underlined}.}
\resizebox{0.96\textwidth}{!}{
\begin{tabular}{l|c|c|c|c|c|c|c|c|c|c}
\toprule[1pt]
\multirow{2}{*}{\textbf{Method}}& \multirow{2}{*}{\textbf{Backbone}}&{\textbf{Transformer}}& \multicolumn{3}{c|}{\textbf{RefCOCO}} & \multicolumn{3}{c|}{\textbf{RefCOCO+}} & \multicolumn{2}{c}{\textbf{G-Ref}} \\ 
\cline{4-11} 
&&{\textbf{block}}& \multicolumn{1}{c|}{\textbf{val}} & \multicolumn{1}{c|}{\textbf{test A}} & \textbf{test B} & \multicolumn{1}{c|}{\textbf{val}} & \multicolumn{1}{c|}{\textbf{test A}} & \textbf{test B} & \multicolumn{1}{c|}{\textbf{val}} & \textbf{test} \\ \hline

RRN \cite{li2018referring} &ResNet101& & \multicolumn{1}{c|}{55.33} & \multicolumn{1}{c|}{57.26} & 53.93 & \multicolumn{1}{c|}{39.75} & \multicolumn{1}{c|}{42.15} & 36.11 & \multicolumn{1}{c|}{-} & - \\
CMSA \cite{ye2019cross} &ResNet101 & &58.32&60.61&55.09&43.76&47.60&37.89&-&-\\
MAttNet \cite{yu2018mattnet} &ResNet101&& \multicolumn{1}{c|}{56.51} & \multicolumn{1}{c|}{62.37} & 51.70 & \multicolumn{1}{c|}{46.67} & \multicolumn{1}{c|}{52.39} & 40.08 & \multicolumn{1}{c|}{47.64} & 48.61 \\
BRINet$^{\dagger}$ \cite{hu2020bi} &ResNet101&& \multicolumn{1}{c|}{60.98} & \multicolumn{1}{c|}{62.99} & 59.21 & \multicolumn{1}{c|}{48.17} & \multicolumn{1}{c|}{52.32} & 42.11 & \multicolumn{1}{c|}{-} & - \\
CMPC \cite{huang2020referring} &ResNet101&& \multicolumn{1}{c|}{61.36} & \multicolumn{1}{c|}{64.53} & 59.64 & \multicolumn{1}{c|}{49.56} & \multicolumn{1}{c|}{53.44} & 43.23 & \multicolumn{1}{c|}{-} & - \\
LSCM$^{\dagger}$ \cite{hui2020linguistic} & ResNet101&&61.47&64.99&59.55&49.34&53.12&43.50&-&-\\
MCN \cite{luo2020multi} &DarkNet53&& \multicolumn{1}{c|}{62.44} & \multicolumn{1}{c|}{64.20} & 59.71 & \multicolumn{1}{c|}{50.62} & \multicolumn{1}{c|}{54.99} & 44.69 & \multicolumn{1}{c|}{49.22} & 49.40 \\
EFN$^{\dagger}$ \cite{feng2021encoder} &ResNet101&& \multicolumn{1}{c|}{62.76} & \multicolumn{1}{c|}{65.69} & 59.67 & \multicolumn{1}{c|}{51.50} & \multicolumn{1}{c|}{55.24} & 43.01 & \multicolumn{1}{c|}{-} & - \\
BUSNet$^{\dagger}$ \cite{yang2021bottom} &ResNet101&&63.27&66.41&61.39&51.76&56.87&44.13&-&- \\
MRLN \cite{hua2023multiple}&DarkNet53&&63.62& 65.57& 60.50 &53.59& 56.78& 47.15& 51.23& 51.02\\
ISFP$^{\dagger}$ \cite{liu2022instance}&DarkNet53&&65.19 &68.45& 62.73& 52.70& 56.77& 46.39& 52.67& 53.00\\
LTS$^{\dagger}$ \cite{jing2021locate} &DarkNet53&\cmark& \multicolumn{1}{c|}{65.43} & \multicolumn{1}{c|}{67.76} & 63.08 & \multicolumn{1}{c|}{54.21} & \multicolumn{1}{c|}{58.32} & 48.02 & \multicolumn{1}{c|}{54.40} & 54.25 \\
SeqTR \cite{zhu2022seqtr}&DarkNet53&\cmark& 67.26& 69.79& 64.12& 54.14& 58.93& 48.19&  55.67 &55.64\\
VLT \cite{ding2021vision} &DarkNet53&\cmark& \multicolumn{1}{c|}{65.65} & \multicolumn{1}{c|}{68.29} & 62.73 & \multicolumn{1}{c|}{55.50} & \multicolumn{1}{c|}{59.20} & 49.36 & \multicolumn{1}{c|}{52.99} & 56.65 \\
ReSTR$^{\dagger}$ \cite{kim2022restr} &ViT-B&\cmark& \multicolumn{1}{c|}{67.22} & \multicolumn{1}{c|}{69.30} & 64.45 & \multicolumn{1}{c|}{55.78} & \multicolumn{1}{c|}{60.44} & 48.27 & \multicolumn{1}{c|}{-} & - \\
CRIS$^{\dagger}$ \cite{wang2022cris} &CLIP-R101&\cmark& \multicolumn{1}{c|}{70.47} & \multicolumn{1}{c|}{73.18} & 66.10 & \multicolumn{1}{c|}{\underline{62.27}} & \multicolumn{1}{c|}{68.08} & 53.60 & \multicolumn{1}{c|}{59.87} & 60.36 \\
LAVT \cite{yang2022lavt} &Swin-B&\cmark& \multicolumn{1}{c|}{\underline{72.73}} & \multicolumn{1}{c|}{\underline{75.82}} & \underline{68.79} & \multicolumn{1}{c|}{62.14} & \multicolumn{1}{c|}{\underline{68.38}} & \underline{55.10} & \multicolumn{1}{c|}{\underline{61.24}} & \underline{62.09} \\ \midrule
CrossVLT (Ours)&ResNet101&\cmark&{70.03}&{73.83}&{64.88}&{59.42}&{64.83}&{49.69}&{59.20}&{60.15}\\
\textbf{CrossVLT (Ours)} &Swin-B& \cmark & \multicolumn{1}{c|}{\textbf{73.44}} & \multicolumn{1}{c|}{\textbf{76.16}} & \textbf{70.15} & \multicolumn{1}{c|}{\textbf{63.60}} & \multicolumn{1}{c|}{\textbf{69.10}} & \textbf{55.23} & \multicolumn{1}{c|}{\textbf{62.68}} &\textbf{63.75} \\
\bottomrule[1pt]
\end{tabular}}
\label{tab:table1}
\end{table*}

\section{Experiments}
\subsection{Datasets}
\textbf{RefCOCO \& RefCOCO+}. RefCOCO \cite{yu2016modeling} and RefCOCO+ \cite{yu2016modeling} are widely used datasets for referring segmentation and were collected from MSCOCO dataset. RefCOCO contains 19,994 images with 142,209 language expressions for 50,000 objects, and RefCOCO+ contains 19,992 images with 141,564 expressions for 49,856 objects. Each expression in RefCOCO and RefCOCO+ contains 3.5 words on average and each images contains 3.9 objects of the same category on average. Expressions in RefCOCO+ do not contain words about absolute locations, and thus, it is more challenging than RefCOCO.

\textbf{G-Ref}. G-Ref \cite{mao2016generation} is also widely used for referring segmentation, which was collected from Amazon Mechanical Turk. G-Ref contains 26,711 images with 104,560 expressions for 54,822 objects. Compared to RefCOCO and RefCOCO+, G-Ref has more complex expressions containing 8.4 words on average, and thus, it is a more challenging dataset than RefCOCO and RefCOCO+.

\subsection{Implementation Details}
\textbf{Experimental settings}. Our method was implemented in PyTorch \cite{Alpher02}. The vision encoder was initialized with Swin-B \cite{liu2021swin} pretrained on ImageNet, and the language encoder was initialized using the official pretrained weights of BERT-base \cite{devlin2019bert} (uncased version). The segmentation decoder was randomly initialized. We trained the model for 40 epochs with a batch size of 16 on RTX 3090 GPU. We used the AdamW optimizer with the learning rate of 3e-4 and adopted the polynomial learning rate decay scheduler. Input images were resized to 480 × 480 pixel resolution. The maximum sentence length was set to 21 words including the \verb|[CLS]| token for three datasets. During the inference, post-processing operations were not applied.

\textbf{Evaluation metrics}. Following previous studies, we used the overall intersection-over-union (oIoU), the mean intersection-over-union (mIoU), and precision at 0.5, 0.7, and 0.9 thresholds. The oIoU is the ratio between the total intersection regions and total union regions of all test sets. The mIoU is the average value of IoUs between the prediction mask and the ground truth of all test sets. The precision is the percentage of test sets that have an IoU score higher than the threshold.

\begin{table}[t]
\centering
\renewcommand{\arraystretch}{}\caption{Main ablation study on RefCOCO \textit{val.} set. Align: Feature-based alignment. Fusion: Cross-aware early fusion.}
\resizebox{\columnwidth}{!}{%
\begin{tabular}{c|c|c|c|c|c|c}
\toprule[1pt]
\multicolumn{1}{c|}{\textbf{Align}} &
  \multicolumn{1}{c|}{\textbf{Fusion}} &
  \textbf{P@0.5} &
  \textbf{P@0.7} &
  \textbf{P@0.9} &
  \multicolumn{1}{c|}{\textbf{mIoU}} &
  \multicolumn{1}{c}{\textbf{oIoU}} \\ \midrule
 &  & 81.61 & 72.48 & 33.70  & 72.31 & 70.77 \\
 \cmark &  & 84.44 & 75.77 & 34.16 & 74.04 & 72.06 \\
 &\cmark  & 85.36 & 76.59 & 35.79 & 75.27 & 73.17 \\
\cmark &\cmark  & \textbf{85.82} & \textbf{77.49} & \textbf{35.97} & \textbf{75.48} & \textbf{73.44} \\
\bottomrule[1pt]
\end{tabular}%
} 
\label{tab:table2}
\end{table}

\subsection{Comparison with the State-of-the-Art}
In \Cref{tab:table1}, we evaluated our CrossVLT with previous state-of-the-art methods on three widely used datasets for referring segmentation using the oIoU metric. Our method surpassed other previous methods on all evaluation splits of all datasets. Compared to the state-of-the-art LAVT on the RefCOCO, CrossVLT had improved performance by 0.71\%, 0.34\%, and 1.36\% on each split, respectively. Further, our model outperformed other state-of-the-art methods on the more challenging RefCOCO+. In addition, on the G-Ref that contains the most challenging data pairs, our CrossVLT achieved performance improvements by 1.44\% and 1.66\% on the validation and test sets, respectively. These improvements indicate that CrossVLT effectively aligns and fuses the cross-modal features, and better understand the complex linguistic expressions and images with dense objects than the previous methods.

\begin{figure}[t] 
\centering
\includegraphics[width=1.0\linewidth]{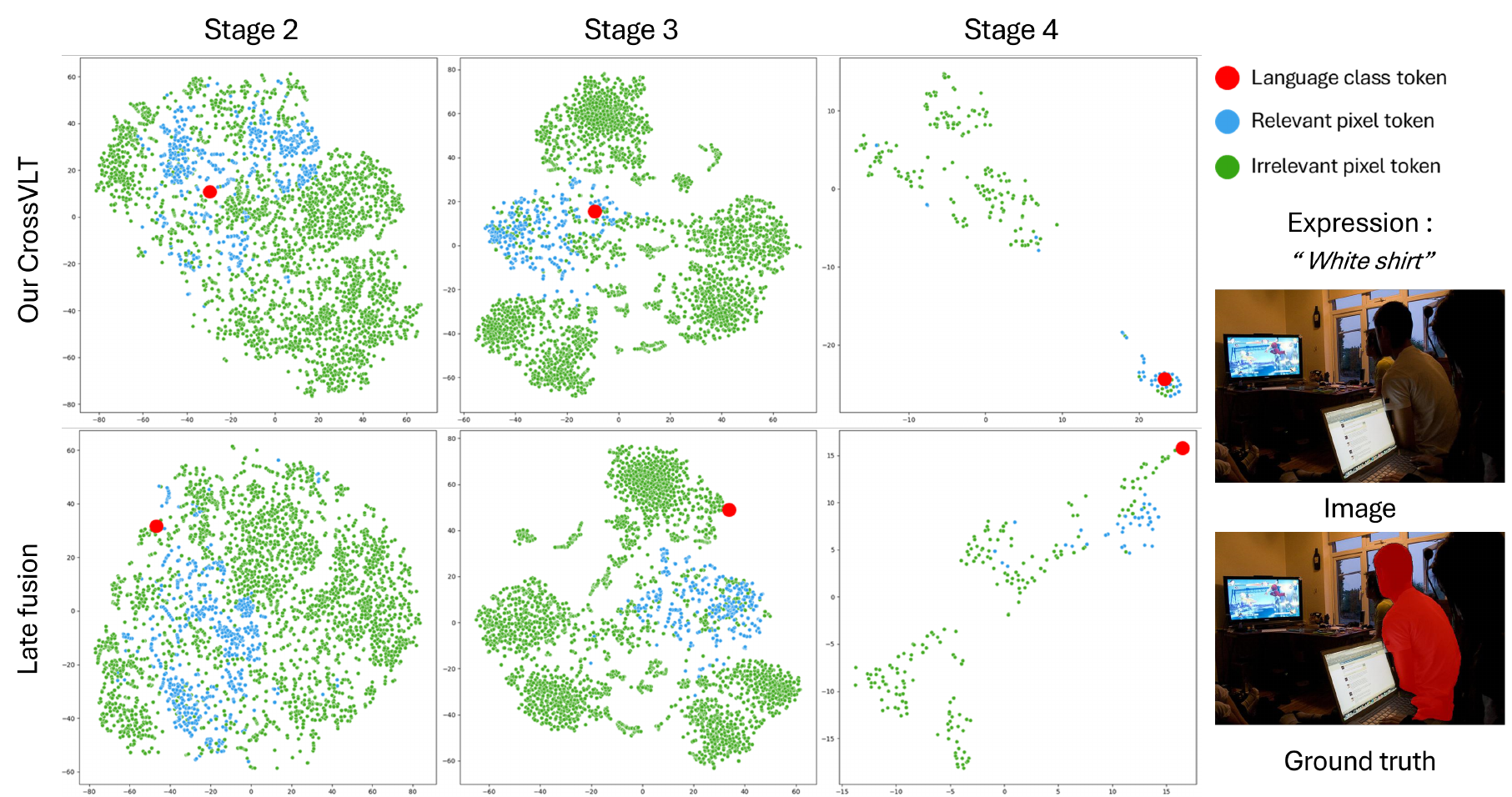}
\caption{Comparison of t-SNE results with the late fusion approach. Red: Language [CLS] token. Blue: Relevant pixel with language expression. Green: Irrelevant pixel.}
\label{fig:fig5}
\end{figure}

\subsection{Ablation Study}

To verify the effectiveness of the main components in our method, we conducted extensive experiments as follows.

\textbf{Cross-aware early fusion and Feature-based alignment.} We analyze the effectiveness of the cross-aware early fusion and the feature-based alignment. In this experiment, the baseline model performs the cross-modal fusion only at the last encoder stage. As displayed in \Cref{tab:table2}, our cross-aware early fusion improves the baseline by 2.96\% and 2.4\% in mIoU and oIoU scores, respectively. In addition, our alignment scheme improves the baseline by 1.73\% and 1.29\% in mIoU and oIoU scores, respectively. Modeling the cross-aware early fusion and feature-based alignment together shows the best performance. These results demonstrate that the cross-aware early fusion and feature-based alignment are effective for modeling the cross-modal context aggregation in referring image segmentation. 

In \cref{fig:fig5}, we visualized t-SNE results of our CrossVLT compared to the late fusion approach. The t-SNE results exhibit the distribution of relevant pixel tokens and irrelevant pixel tokens with language \verb|[CLS]| token in cross-modal embedding space. In t-SNE of our CrossVLT, the relevant pixel tokens are embedded closer to the language token than in the late fusion results by learning the cross-modal interactions. Therefore, this result visually demonstrates the effectiveness of our approach.

\begin{table}[t]
\centering
\renewcommand{\arraystretch}{}\caption{Ablation studies on the validation set of RefCOCO. `Bi' and `Uni' indicate bidirectional and unidirectional fusion, respectively. (w/) : with the feature-based alignment. (w/o) : without the feature-based alignment.}
\resizebox{\columnwidth}{!}{%
\begin{tabular}{l|c|c|c|c|c}
\toprule[1pt]
\textbf{} & \textbf{P@0.5} & \textbf{P@0.7} & \textbf{P@0.9} & \textbf{mIoU} & \textbf{oIoU} \\ \midrule
\multicolumn{6}{l}{(a) Stages that applied only cross-aware early fusion}\\
\midrule
{[}4{]}&81.61& 72.48& 33.70& 72.31& 70.77\\
{[}3, 4{]}&84.47&75.17&34.89&74.25&72.29\\
{[}2, 3, 4{]}&84.99&76.12&35.51&74.93&72.83\\
{[}1, 2, 3, 4{]}&\textbf{85.36}&\textbf{76.59}&\textbf{35.79}&\textbf{75.27}&\textbf{73.17}\\
\midrule
\multicolumn{6}{l}{(b) Stages that applied only feature-based alignment}\\
\midrule
{[}4{]}&83.00&73.35&33.97&73.25&71.58\\
{[}3, 4{]}&83.22&73.56&33.76&73.49&71.69\\
{[}2, 3, 4{]}&83.57&74.79&33.90&73.62&71.84\\
{[}1, 2, 3, 4{]}&\textbf{84.44}&\textbf{75.77}&\textbf{34.16}&\textbf{74.04}&\textbf{72.06}\\
\midrule
\multicolumn{6}{l}{(c) Stages that applied alignment and fusion together} \\
\midrule
{[}4{]}  &  83.00  &  73.35  &   33.97  &  73.25 & 71.58 \\
{[}3, 4{]}  & 84.84    & 76.56     & 35.68   & 74.61     & 72.42 \\
{[}2, 3, 4{]}  &     85.37     &  76.90        &      35.92     &    74.99    & 72.82 \\
{[}1, 2, 3, 4{]}& \textbf{85.82}& \textbf{77.49}& \textbf{35.97}& \textbf{75.48}& \textbf{73.44}\\
\midrule
\multicolumn{6}{l}{(d) Effectiveness of the cross-aware fusion} \\
\midrule
{Uni (w/o)} & 84.96 & 76.44 & 35.31 & 74.84 & 72.82 \\
{Uni (w/)}  & 85.45 & 77.19 & 35.93 & 75.05 & 73.03 \\
{Bi (w/o)} & 85.36 & 76.59 & 35.76 & 75.28 & 73.17 \\
{Bi (w/)}  & \textbf{85.82} & \textbf{77.49} & \textbf{35.97} & \textbf{75.48} & \textbf{73.44} \\
\midrule
\multicolumn{6}{l}{(e) Applying different loss to all intermediate encoder stages} \\
\midrule
Auxiliary loss & 84.89 &76.36&35.66&74.96&72.62\\
Alignment loss & \textbf{85.82} & \textbf{77.49} & \textbf{35.97} & \textbf{75.48} & \textbf{73.44} \\
\bottomrule[1pt]
\end{tabular}%
}
\label{tab:table3}
\end{table}

In \Cref{tab:table3} (a), we experimented with applying the cross-aware early fusion to each encoder stage incrementally. Applying the cross-aware early fusion on all stage led to significant improvements of 2.96\% and 2.4\% in mIoU and oIoU scores. These results indicate that cross-aware early fusion with the stage-divided both encoders enables the exchange of abundant information between the vision and language encoders. In \Cref{tab:table3} (b), we experimented with applying the feature-based alignment to each encoder stage incrementally. Applying the feature-based alignment on all stages also improved performance by 0.79\% and 0.48\% in mIoU and oIoU scores. These results indicate that our scheme helps to align the intermediate cross-modal features for effective fusion.

\begin{table}[t]
\centering
\renewcommand{\arraystretch}{}\caption{Comparison of the performance on the number of layers in each stage of the language encoder. Models are trained without pretrained weights.}
\resizebox{\columnwidth}{!}{
\begin{tabular}{c|c|c|c|c|c}
\toprule[1pt]
\textbf{Number of layers} & \textbf{P@0.5} & \textbf{P@0.7} & \textbf{P@0.9} & \textbf{mIoU} & \textbf{oIoU} \\
\midrule
{[2, 2, 2, 6]} & 84.24 & 75.56 & 34.63 & 74.04 & 72.03 \\
{[3, 3, 3, 3]} & 84.53 & 76.17 & 35.51 & 74.39 & 72.09 \\
{[6, 2, 2, 2]} & \textbf{84.91} & \textbf{77.09} & \textbf{36.43} & \textbf{74.90} &\textbf{72.52} \\
{[8, 2, 1, 1]} & 84.46 & 76.43 & 36.25 & 74.50 & 72.19 \\
\bottomrule[1pt]
\end{tabular}
}
\label{tab:tableA1}
\end{table}

\begin{table}[t]
\centering
\renewcommand{\arraystretch}{1}\caption{Performance comparison under fair conditions using the same backbone and language encoder.}
\resizebox{\columnwidth}{!}{%
\begin{tabular}{l|c|c|c|c|c}
\toprule[1pt]
\textbf{Method} & \textbf{P@0.5} & \textbf{P@0.7} & \textbf{P@0.9} & \textbf{mIoU} & \textbf{oIoU} \\
\midrule
EFN \cite{feng2021encoder} & 82.55 & 73.27 & 31.68 & 72.95 & 70.76 \\
VLT \cite{ding2021vision} & 83.24 & 72.81 & 24.64 & 71.98 & 70.89 \\
LAVT \cite{yang2022lavt} & 84.46 & 75.28 & 34.30 & 74.46 & 72.73 \\ 
\midrule
\textbf{Ours} & \textbf{85.82} & \textbf{77.49} & \textbf{35.97} & \textbf{75.48} & \textbf{73.44} \\
\bottomrule[1pt]
\end{tabular}
}

\label{tab:tableA3}
\end{table}

\begin{table}[t]
\centering
\renewcommand{\arraystretch}{}\caption{Ablation study of using different baseline methods on RefCOCO \textit{val.} set. All experiments use Swin-B as the visual backbone and BERT-base as the language encoder.}
\resizebox{\columnwidth}{!}{%
\begin{tabular}{l|c|c|c|c|c}
\toprule[1pt]
  \multicolumn{1}{c|}{{\textbf{Method}}} &
  {\textbf{P@0.5}} &
  {\textbf{P@0.7}} &
  {\textbf{P@0.9}} &
  \multicolumn{1}{c|}{{\textbf{mIoU}}} &
  \multicolumn{1}{c}{{\textbf{oIoU}}} \\ \midrule
 {VLT \cite{ding2021vision}}  & {83.24} & {72.81} & {24.64} & {71.98} & {70.89}\\
 {Ours + VLT \cite{ding2021vision}} &  {\textbf{85.73}} & {\textbf{77.81}} & {\textbf{35.56}} & {\textbf{75.33}} & {\textbf{73.60}}   \\
 \midrule
 {CGFormer \cite{tang2023contrastive}}  & {87.23} & {78.69} & {38.77} & {76.93} & {74.75} \\
 {Ours + CGFormer \cite{tang2023contrastive}} &{\textbf{88.17}} & {\textbf{79.90}} & {\textbf{39.88}} & {\textbf{77.75}} & {\textbf{75.63}} \\
\bottomrule[1pt]
\end{tabular}%
} 
\label{tab:new}
\end{table}

\begin{figure}[t]
\centering
\includegraphics[width=0.8\columnwidth]{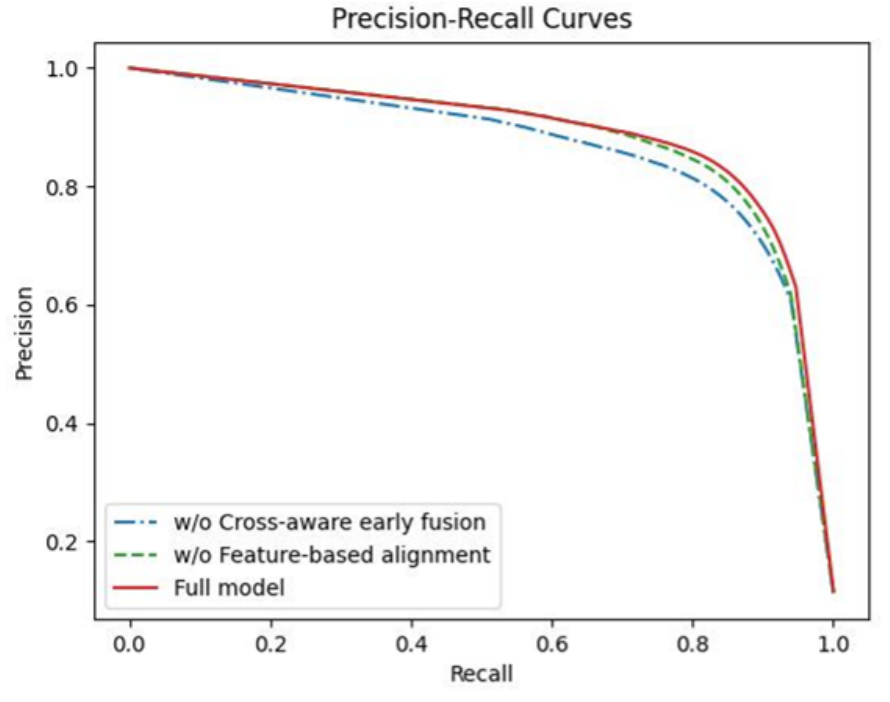} 
\caption{{Precision-Recall (PR) curves of our model and two ablation models on RefCOCO validation set. }}
\label{fig:curve}
\end{figure}

\begin{figure}[t]
\centering
\includegraphics[width=\columnwidth]{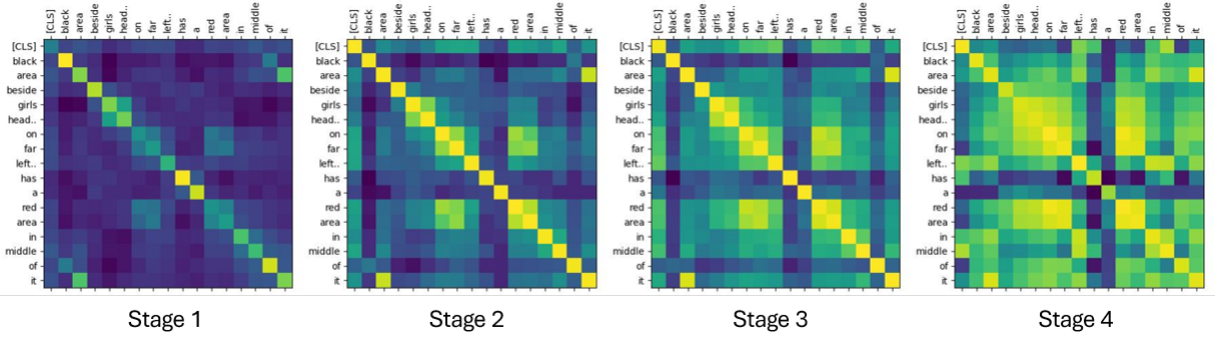}
\caption{Visualizing contextual similarity of all word features in each stage. }
\label{fig:fig6}
\end{figure}

\textbf{Importance of applying the alignment and fusion together in every stage.} In \Cref{tab:table3} (c), we evaluated the effectiveness of applying the alignment and fusion together in every stage. The performance improved as the number of stages to which alignment and fusion are applied increased. That is, the best performance occurred when they are applied at all stages. Compared with the setting of [4], the mIoU and oIoU scores at the setting of [1, 2, 3, 4] increased by 2.23\% and 1.86\%, respectively. We take this as evidence that applying the fusion and alignment together at every stage is effective by considering the low-level to high-level information of the cross-modalities.

\textbf{Importance of the cross-aware fusion.} In \Cref{tab:table3} (d), all models performed the early fusion to exclude the effectiveness of the early fusion and to purely verify the effectiveness of the cross-aware fusion. The `Uni' fusion (\textit{i.e.} vision-only early fusion), where only vision stages perform the early fusion with the intermediate features of the language stage, showed the degraded performance compared with the `Bi' fusion models (\textit{i.e.} cross-aware early fusion). The results indicate that the cross-aware fusion enables to capture the rich contextual information by considering each other’s perspectives.

\textbf{Suitability of alignment loss.} We experimented on replacing our loss with common segmentation auxiliary loss to help uncover the role of alignment loss. In \Cref{tab:table3} (e), applying alignment loss perform much better, whereas applying common auxiliary loss on all stages rather causes conflict in learning the encoder features and degrades performance. This indicates that the design of our alignment loss is more suitable for applying to intermediate encoder features and beneficial for the cross-modal alignments.

\textbf{Optimal settings for the stage-divided language encoder.} In the language encoder, determining the criteria for dividing the stages is unclear. We empirically found the optimal hyper parameters for designing an effective stage-divided language encoder. We experimented with the number of layers in each stage by limiting the total number of layers in the language encoder to 12 based on the BERT-base model \cite{devlin2019bert}. As shown in \Cref{tab:tableA1}, the setting [6, 2, 2, 2] performed the best under the same conditions without using pretrained weights. The setting [2, 2, 2, 6] cannot fully extract the low-level linguistic information. The setting [8, 2, 1, 1] is unsuitable for utilizing the high-level semantic information. Therefore, we adopted the optimal setting [6, 2, 2, 2].

\textbf{Fair comparison using the same backbone.} In \Cref{tab:tableA3}, we compared the proposed model with the other state-of-the-art models fairly, using the same vision and language encoders on all models. Swin-B \cite{liu2021swin} and BERT-base \cite{devlin2019bert} were used as the vision and language encoders, respectively. Our CrossVLT surpassed the previous models under the same conditions. This result demonstrates that the performance improvement of CrossVLT was not the effect of the particular backbone and language encoder (i.e., Swin-B \cite{liu2021swin} and BERT-base \cite{devlin2019bert}).

\textbf{Comparison with different baseline methods.} In Table \ref{tab:new}, we applied our method to different state-of-the-art methods (\textit{i.e.}, VLT and CGFormer) that consist of more complicated cross-modal decoder. As shown in Table \ref{tab:new}, both models combined with our method showed significant improvements in oIoU performance of 2.71\% and 0.88\% compared to VLT and CGFormer, respectively. These results indicate that our method can be applied to other methods and leads to additional performance gains for other methods.

\textbf{Precision-recall analysis.} In Fig. \ref{fig:curve}, we analyzed the precision-recall (PR) curves of our model and the ablation models. The area under the PR curve (AUC-PR) summarizes the overall performance of the model across different threshold values. A higher AUC-PR indicates a better-performing model. As shown in Fig. \ref{fig:curve}, our full model (red solid) had the highest AUC-PR compared to other ablation models. From around 0.7 recall, our full model maintained its advantage in precision over the alignment ablation model (green dashed).

\subsection{Visualizations}
\textbf{Visualizing language features.} Fig.\ref{fig:fig6} showed that each stage of the language encoder captures different information. The lower stage contains the syntactic information and the higher stage captures long-range relations. Our method exploits this abundant information of the intermediate features for the cross-modal fusion and alignment.

\begin{figure*}[t!]
\centering{\includegraphics[width=\textwidth]{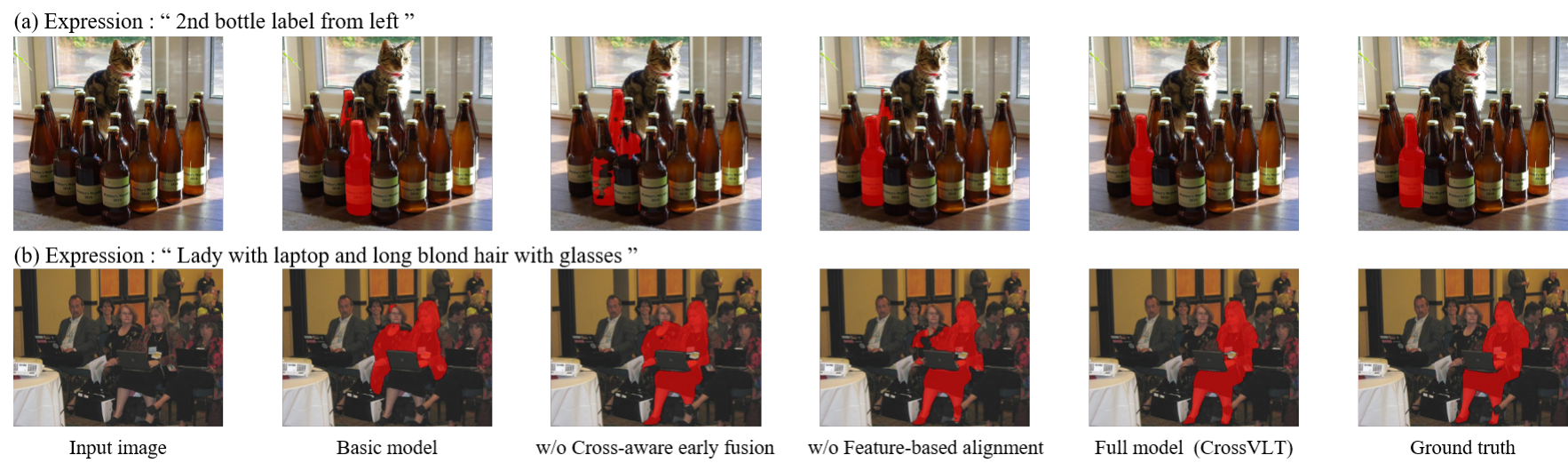}}
\caption{Visualization predictions of our full model and other ablated models on the test set of RefCOCO. Basic model represents a ablated model removed the cross-aware fusion and feature-based alignment. }
\label{fig:fig9}
\end{figure*}

\begin{figure*}[t] 
\centering\includegraphics[width=\linewidth]{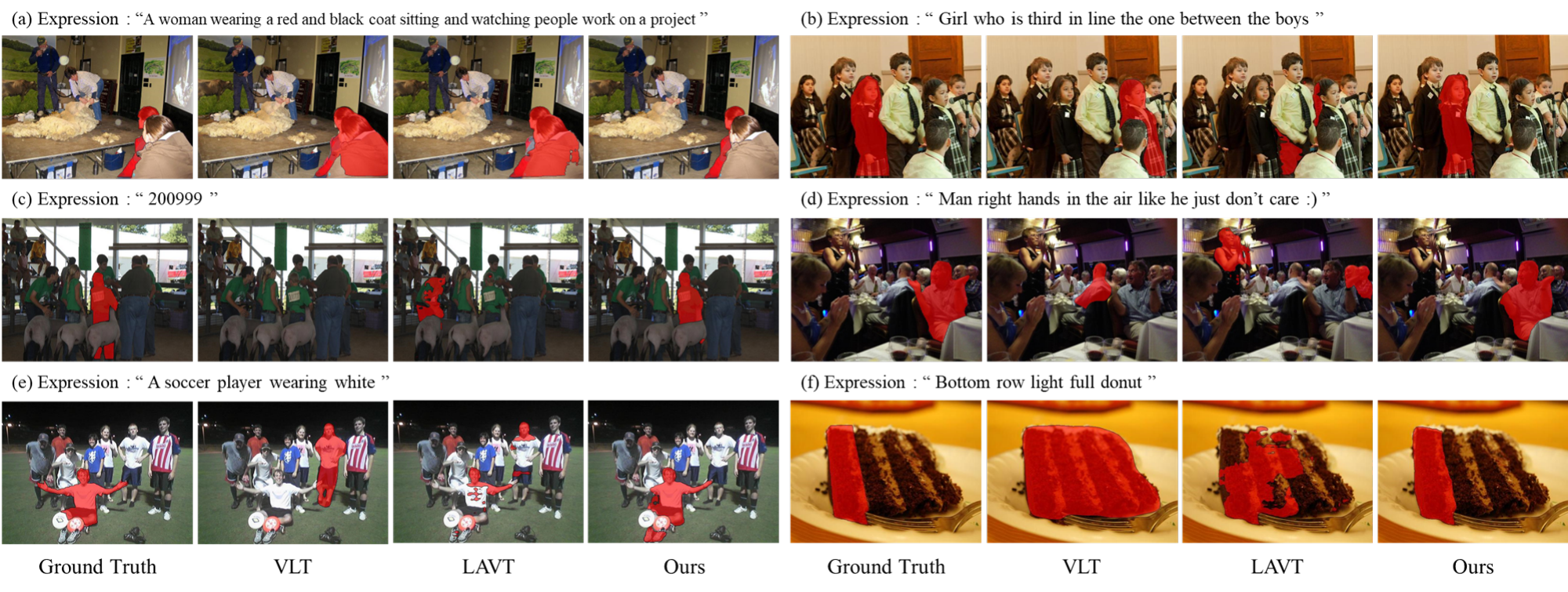}
\caption{Qualitative results compared with the proposed and previous state-of-the-art methods on different types of images and language expressions. To understand the expression of (c), the back number of the target object is ``200999".}
\label{fig:fig7}
\end{figure*}

\textbf{Qualitative results.}
In \cref{fig:fig9}, we present qualitative results of our full model and other ablated models (\textit{i.e.} without cross-aware early fusion, without feature-based alignment and without both components) to demonstrate the effectiveness of each component. As shown in examples of \cref{fig:fig9}, our full model segmented the target regions more elaborately than the other ablated models. These qualitative results indicate that applying both the feature-based alignment and cross-aware early fusion is an effective approach for referring image segmentation.

In \cref{fig:fig7}, we visualized the segmentation results of our method and the two state-of-the-art methods: the late fusion model \cite{ding2021vision} and the vision-only early fusion model \cite{yang2022lavt}. For \cref{fig:fig7} (a) and (b), which include the complex expressions and complicated images with dense objects, our method highlighted the target regions more accurately than the previous models. For the other four examples that include ambiguous expressions, our method correctly segmented the targets corresponding to the expressions, whereas other methods incorrectly predicted the objects and uncertainly segmented the regions. These results indicate that CrossVLT can better understand the context of images and expressions containing complicated relationships.

\begin{figure*}[t!]
\centering{\includegraphics[width=\textwidth]{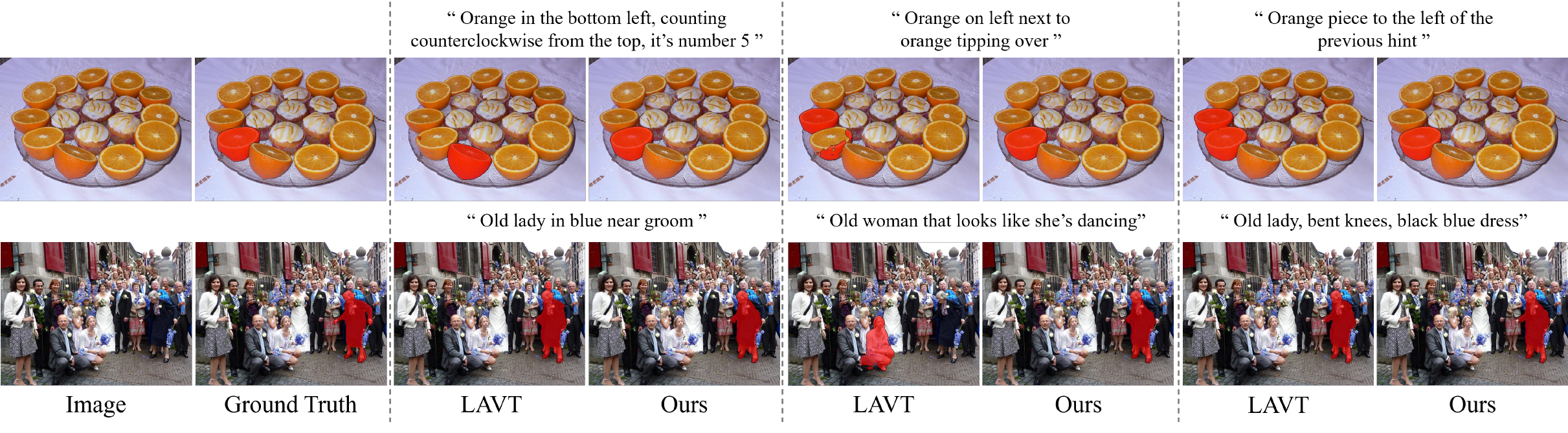}}
\caption{Visualization two examples of our method and the previous state-of-the-art method (LAVT \cite{yang2022lavt}) on various language expressions that describe the same object in the image.}
\label{fig:fig8}
\end{figure*}

\begin{figure*}[t] 
\centering
\includegraphics[width=\linewidth]{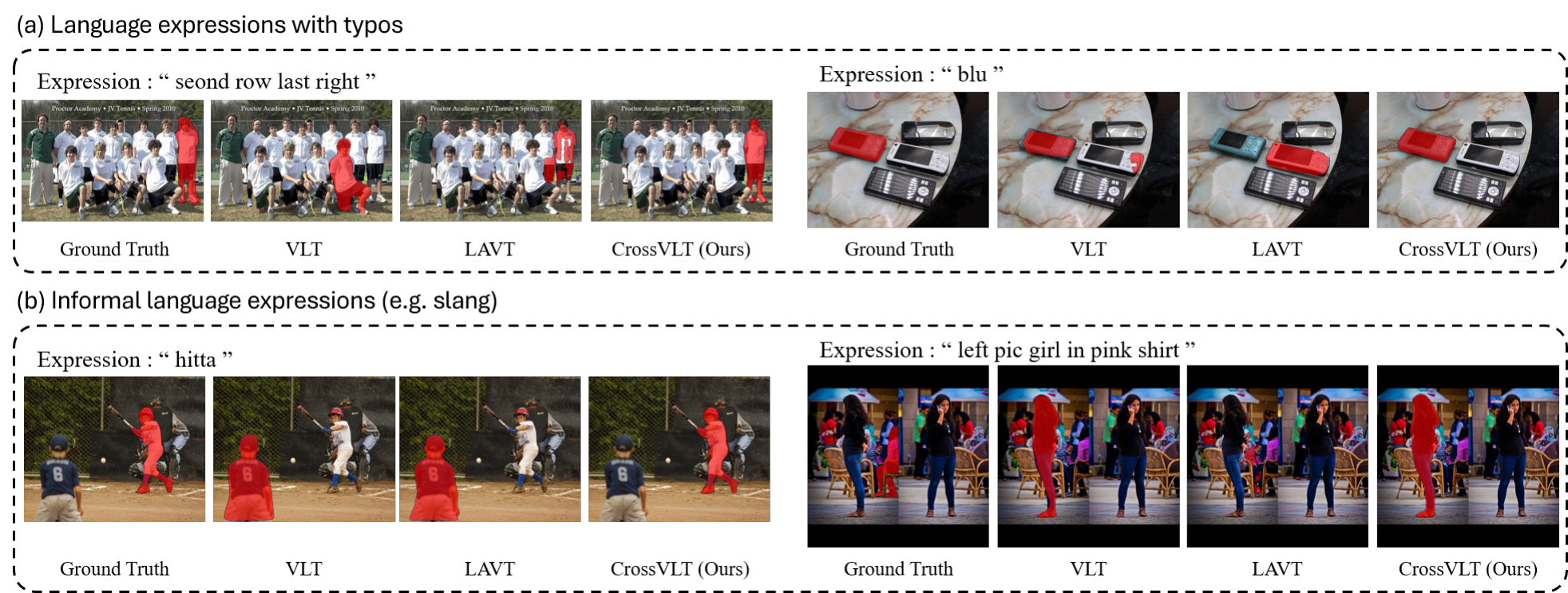}
\caption{Visualizations of our method and the previous state-of-the-art methods on challenging types of linguistic expressions.}
\label{fig:fig10}
\end{figure*}

In \cref{fig:fig8}, given the different expressions that describe the same target, we consistently predicted the target, whereas the other method inconsistently predicted the object. These results verify that our method enables dealing with various language expressions by enhancing the robustness of both encoders.

\textbf{The Robustness of CrossVLT.}
As shown in \cref{fig:fig10} (a), when previous methods encountered language expressions with typos (e.g. ``seond" and ``blu"), they are confused in understanding the meaning of expressions. Additionally, as shown in \cref{fig:fig10} (b), informal language expressions (e.g. ``hitta" and ``pic") also made it difficult for the network to capture the context of the expressions in previous methods. Given these challenging types of expressions, our CrossVLT, unlike other state-of-the-art models, correctly determined the target regions by referring to the visual perspectives in understanding the meaning of the expressions. These results indicate that our method enhances the robustness of each encoder and the ability of understanding the contexts.

\section{Conclusion}
This paper proposed a novel network for referring image segmentation, Cross-aware Early Fusion with Stage-divided Vision and Language Transformer Encoders (CrossVLT), which leverages cross-modal features alternately traversing through each stage of the two transformer encoders to mutually enhance the robustness of each encoder. We also introduced the feature-based alignment scheme that performs contrastive learning using the features of the intermediate levels in each encoder to enhance the capability of aligning the cross-modal features. Experiments demonstrated the effectiveness of our method on three public datasets. We hope our method can motivate further research for various multi-modal tasks and the feature-based alignment scheme can be applied by designing the alignment loss to suit their tasks.

\section*{Acknowledgments}
This research was supported by by Samsung Electronics under Grant IO201218-08232-01, the MSIT (Ministry of Science and ICT), Korea, under the ITRC (Information Technology Research Center) support program (IITP-2023-RS-2023-00260091) supervised by the IITP (Institute for Information \& Communications Technology Planning \& Evaluation), by the National Research Foundation of Korea (NRF) grant funded by the Korea government (MSIT) (No. 2021R1A2C1004208), and by the National Research Foundation of Korea (NRF) grant funded by the Korea government (MSIT) (No. 2020M3H4A1A02084899).

\bibliographystyle{IEEEtran}
\bibliography{IEEEabrv, reference}
\newpage
\begin{IEEEbiography}[{\includegraphics[width=1in,height=1.25in,clip,keepaspectratio]{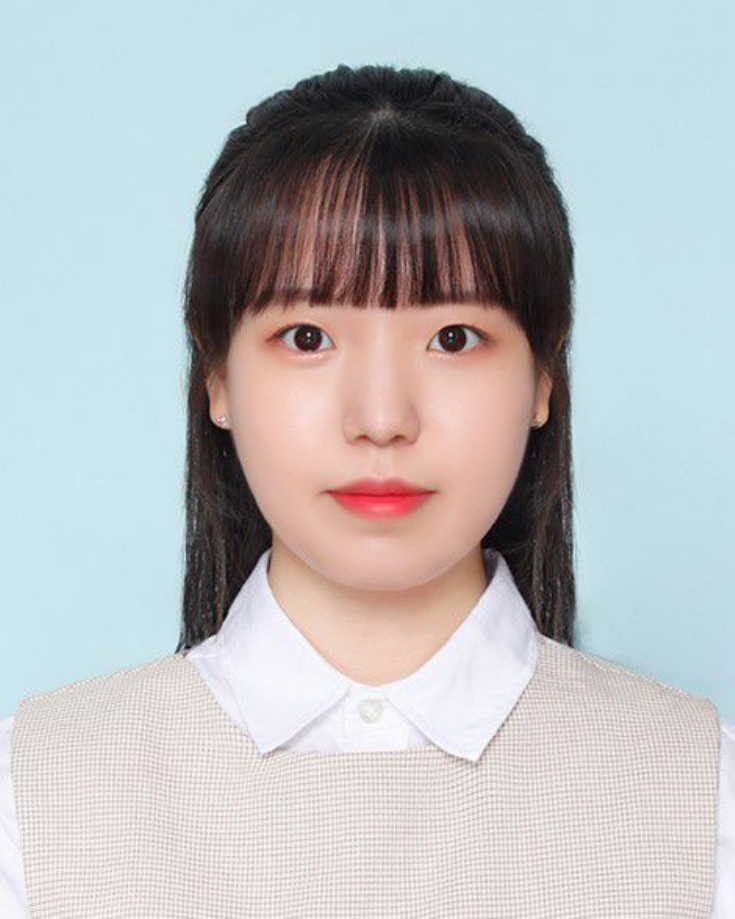}}]{\textbf{Yubin Cho}} received the B.S. degree with double major in Mechanical Engineering and Artificial Intelligence from Sogang University, South Korea, in 2022. She is currently working toward the M.S. degree with the School of Artificial Intelligence, Sogang University. Her current research interests include computer vision, multi-modal learning and deep learning.
\end{IEEEbiography}

{\vskip -1\baselineskip plus -1fil}

\begin{IEEEbiography}[{\includegraphics[width=1in,height=1.25in,clip,keepaspectratio]{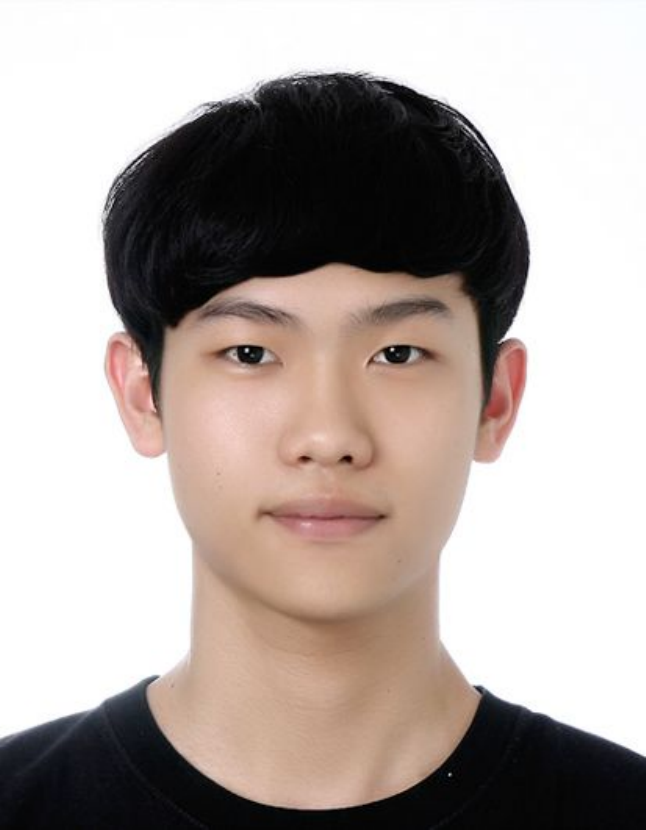}}]{\textbf{Hyunwoo Yu}} received the B.S. degree in Physics from Kangwon University, South Korea, in 2021. He is currently working toward the Ph.D. degree with the School of Electronic Engineering, Sogang University, South Korea. His current research interests include computer vision, multi-modal learning and pixel-level scene understanding.
\end{IEEEbiography}

{\vskip -1\baselineskip plus -1fil}

\begin{IEEEbiography}[{\includegraphics[width=1in,height=1.25in,clip,keepaspectratio]{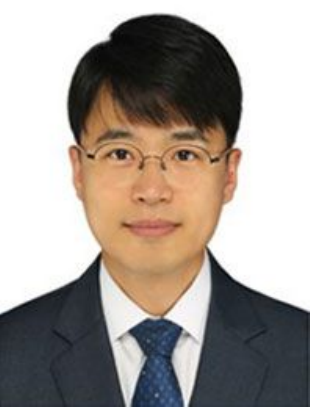}}]{\textbf{Suk-ju Kang}}
(Member, IEEE) received the B.S. degree in electronic engineering from Sogang University, Seoul, South Korea, in 2006, and the Ph.D. degree in electrical and computer engineering from the Pohang University of Science
and Technology, Pohang, South Korea, in 2011. From 2011 to 2012, he was a Senior Researcher with LG Display Co., Ltd., Seoul, where he was a Project Leader for resolution enhancement and multiview 3-D system projects. From 2012 to 2015, he was an Assistant Professor of Electrical Engineering with Dong-A University, Busan, South Korea. He is currently a Professor of Electronic Engineering with Sogang University, Seoul. His current research interests include  computer vision, image analysis and enhancement, video processing, multimedia signal processing, digital system design, and deep learning. He was a recipient of the IEIE/IEEE Joint Award for Young IT Engineer of the Year in 2019 and the Merck Young Scientist Award in 2022. He served as an Associate Editor for IEEE Transactions on Circuits and Systems for Video Technology from 2023. 
\end{IEEEbiography}


\newpage

\vfill

\clearpage

\appendix
\appendices
\begin{itemize}
    \item In Appendix \ref{app_a}, we provide the detailed quantitative results of our model on three public referring image segmentation datasets.
    \item In Appendix \ref{app_b}, we provide the visualizations of our segmentation results, which are better than the ground truth.
    \item In Appendix \ref{app_c}, we provide additional visualizations on various types of language expressions.
\end{itemize}

\section{Detailed quantitative results}\label{app_a}

\begin{table*}[t]
\centering \caption{Precision, mean IoU (\%) and overall IoU(\%) of the proposed method on three standard benchmark datasets.}
\resizebox{0.8\linewidth}{!}{
\begin{tabular}{c|c|c|c|c|c|c|c|c}
\toprule[1pt]
\multicolumn{2}{c|}{\textbf{Dataset}}&\textbf{P@0.5}&\textbf{P@0.6}&\textbf{P@0.7}&\textbf{P@0.8}&\textbf{P@0.9}&\textbf{mIoU}&\textbf{oIoU} \\ 
\midrule
\multirow{3}{*}{RefCOCO}&val&85.82&	82.80&	77.49&	67.21&	35.97&	75.48&	73.44\\
&test A&88.92&	86.25&	81.62&	69.79&	36.68&	77.54&	76.16\\
&test B&81.35&	77.74&	72.27&	62.63&	37.64&	72.69&	70.15\\
\midrule
\multirow{3}{*}{RefCOCO+}&val&76.41&	73.09&	68.49&	59.09&	31.39&	67.27&	63.60\\
&test A&82.43&79.57	&74.80&	64.65&	32.89&	72.00	&69.10\\
&test B&67.19&	63.31&	58.89&	49.93&	29.15&	60.09&	55.23\\
\midrule
\multirow{2}{*}{G-Ref}&val&74.75&70.45	&64.58&	54.21&	28.57&	66.21&	62.68\\
&test&71.54&	66.38&	59.00&	48.21&	23.10&	62.09&	63.75\\
\bottomrule[1pt]
\end{tabular}}
\label{tab:tableB}
\end{table*}

\Cref{tab:tableB} shows the detailed precision, mIoU(\%) and oIoU(\%) scores of our model on RefCOCO, RefCOCO+, and G-Ref datasets to complement the quantitative results in Table \uppercase\expandafter{\romannumeral1} of the main paper.

\section{Visualization of Predictions Better than Ground Truth}\label{app_b}
\begin{figure}[t] 
\centering
\includegraphics[width=\columnwidth]{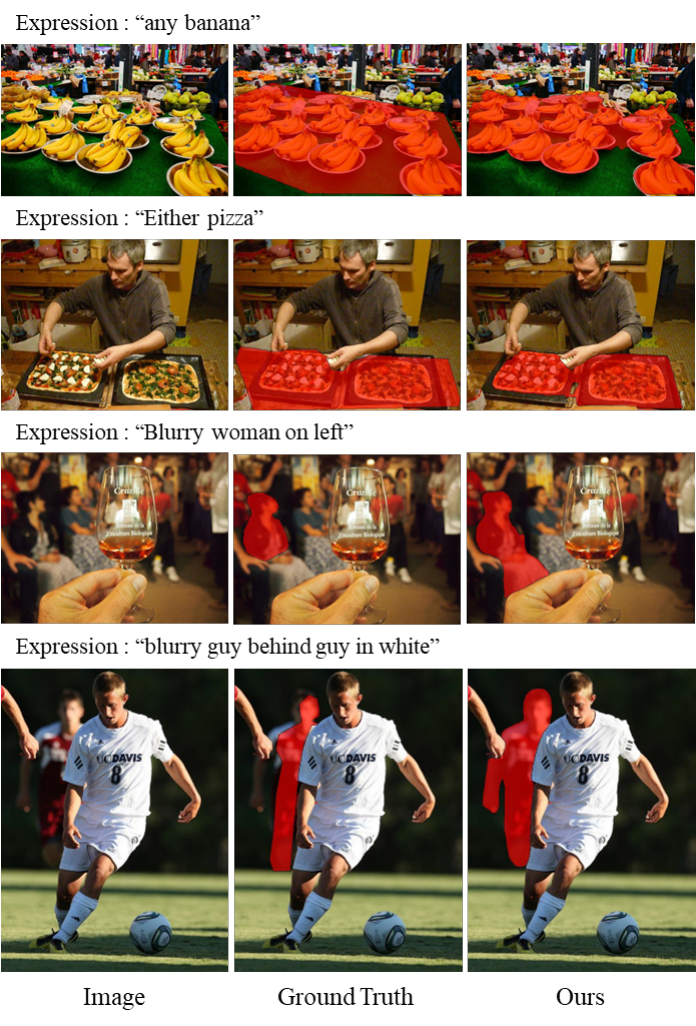}
\caption{Visualization of our predictions that include the detailed regions of the target object more accurately than the ground truth. }
\label{fig:fig15}
\end{figure}

In \cref{fig:fig15}, we visualized our predictions in which the target regions were segmented more precisely than the ground truth. In the ground truth of the first and second rows, the annotations included non-target regions (e.g. part of the table). In the ground truth of the third and fourth rows, the annotations excluded the target regions (e.g. part of the body). However, our predictions included the detailed regions of target objects accurately.

\section{Visualization of Additional Results}\label{app_c}
\begin{figure*}[t] 
\centering
\includegraphics[width=\linewidth]{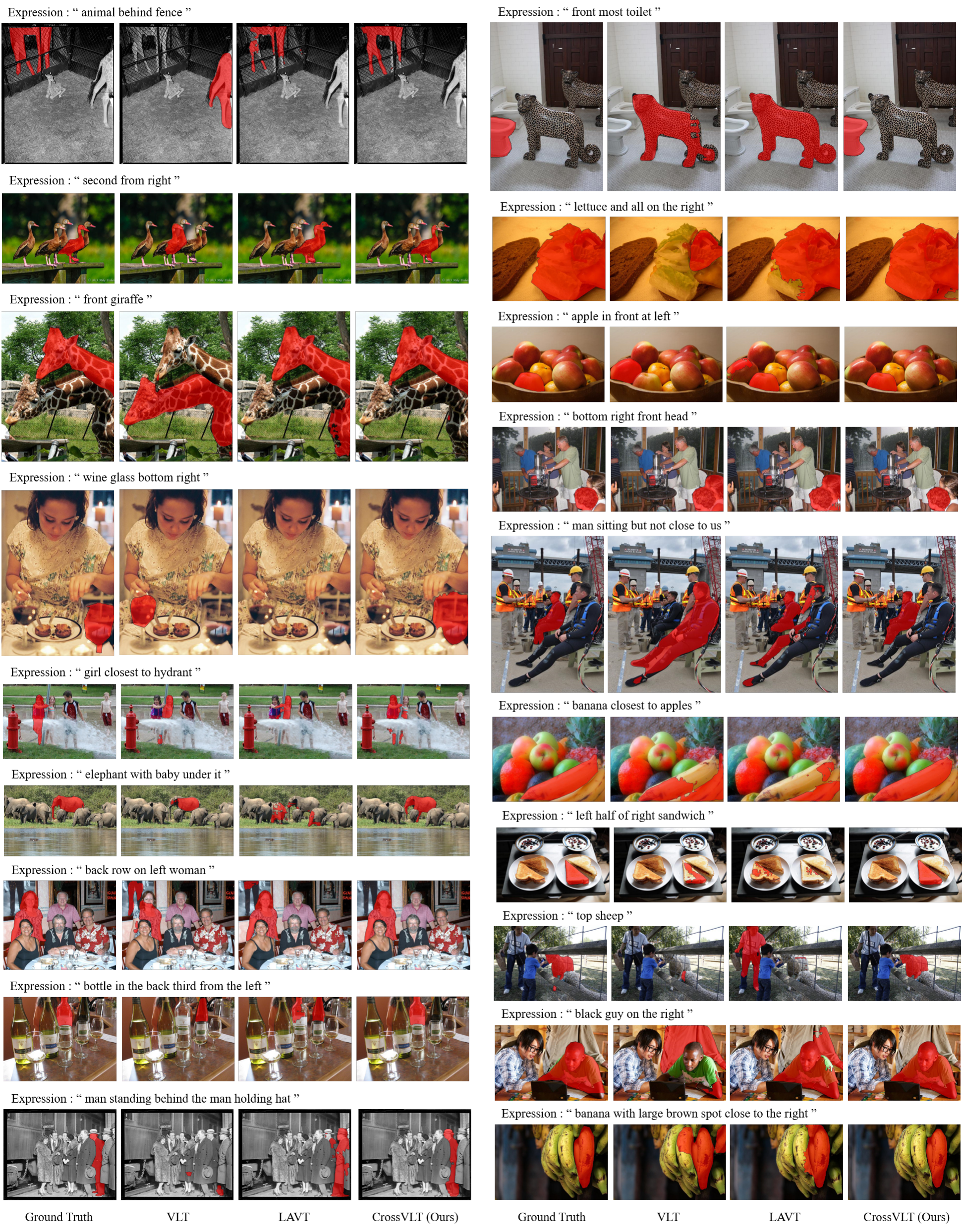}
\caption{Qualitative results of the proposed and previous state-of-the-art methods on various language expressions describing locations. }
\label{fig:fig14}
\end{figure*}

\begin{figure*}[t] 
\centering
\includegraphics[width=\linewidth]{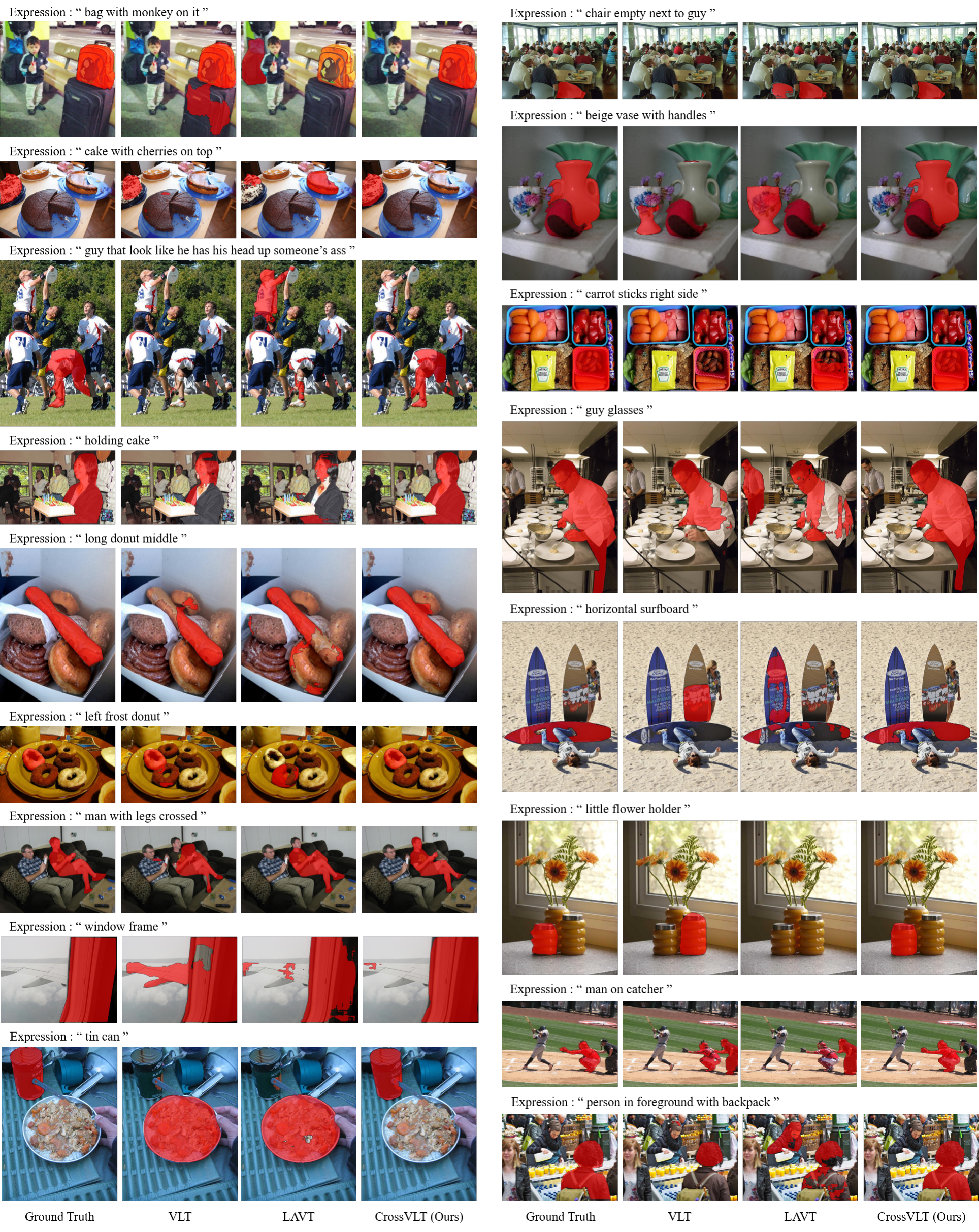}
\caption{Qualitative results of the proposed and previous state-of-the-art methods on various language expressions describing the attribute of the objects.}
\label{fig:fig11}
\end{figure*}

\begin{figure*}[t] 
\centering
\includegraphics[width=\linewidth]{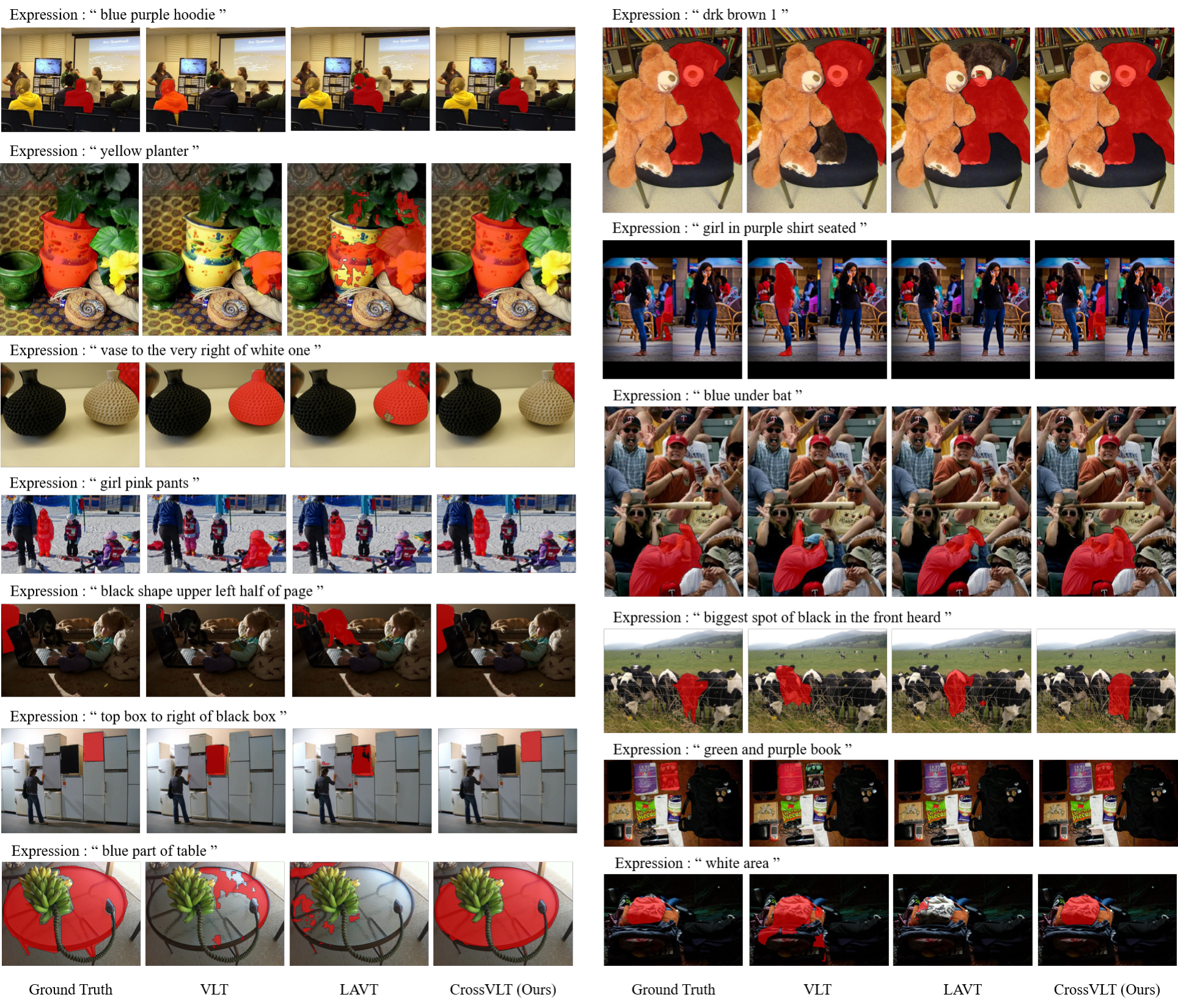}
\caption{Qualitative results of the proposed and previous state-of-the-art methods on various language expressions containing colors.}
\label{fig:fig12}
\end{figure*}

\begin{figure*}[ht] 
\centering
\includegraphics[width=\linewidth]{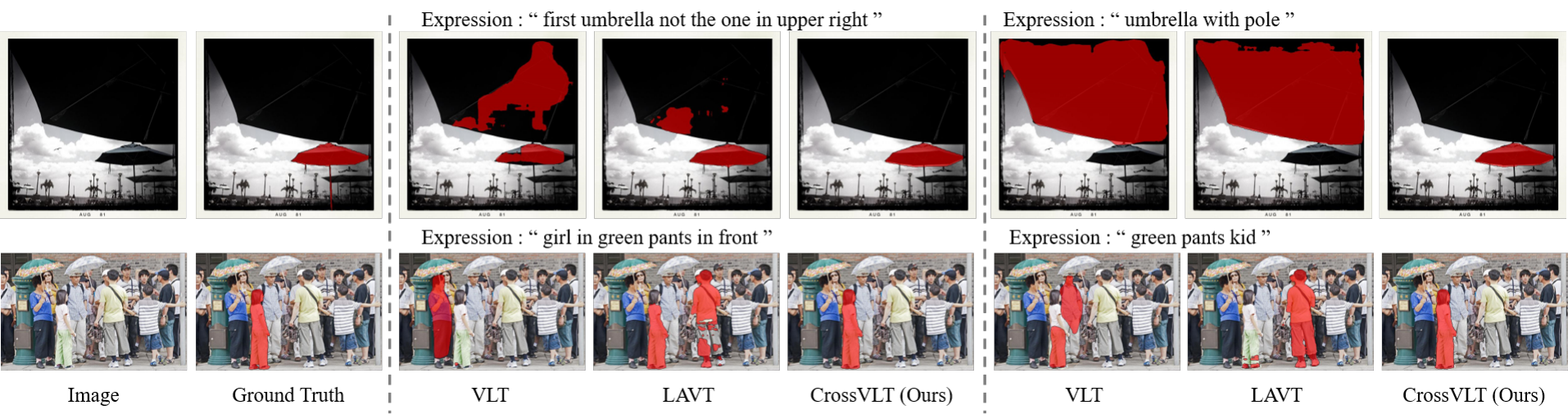}
\caption{Visualization additional examples of our method and previous state-of-the-art methods on various language expressions describing the same object in the image. Our method consistently determines the target regions by robustly dealing with various expressions, whereas other methods inconsistently predict the target regions.}
\label{fig:fig13}
\end{figure*}

 From \cref{fig:fig14,fig:fig11,fig:fig12,fig:fig13}, we visualized the additional qualitative results on various types of language expressions to clearly prove the high level of competence in understanding the context of images and language expressions. As shown in \cref{fig:fig14}, our model could understand expressions describing relative locations between the objects (e.g. ``Man standing behind the man holding hat"). Compared to the previous methods, our method better processed the attribute words (e.g. ``frost", ``horizontal" and ``empty") as displayed in \cref{fig:fig11}. In \cref{fig:fig12}, our method also recognized colors of the objects (e.g. ``white area" and ``blue part of table") more accurately than other methods. Moreover, in \cref{fig:fig13}, we visualized more examples on predicting the same target object described by different language expressions to complement the qualitative results in Fig. 8 of the main paper. Our model consistently determined the target object by robustly dealing with different language expressions, whereas other previous models dealt with them inconsistently.

\end{document}